\documentclass[10pt,twocolumn,letterpaper]{article}

\usepackage{iccv}              % To produce the CAMERA-READY version
\usepackage[accsupp]{axessibility}  % Improves PDF readability for those with disabilities.

\definecolor{cc1}{rgb}{1.0, 0.44, 0.37}
\definecolor{cc2}{rgb}{0.0, 0.2, 0.6}
\definecolor{cc3}{RGB}{255, 191, 0}
\definecolor{cc4}{RGB}{0, 128, 128}
\definecolor{latency}{RGB}{82, 142, 249}
\definecolor{gmacs}{RGB}{255, 68, 68}
\definecolor{softmax}{RGB}{252, 168, 75}
\definecolor{linear}{RGB}{132, 176, 113}

\usepackage{booktabs}
\usepackage{multicol}
\usepackage{multirow}
\usepackage{color}
\usepackage{caption}
\usepackage{hhline}
\usepackage{pifont}
\usepackage{threeparttable}
\usepackage{makecell}
\usepackage{algorithm} 
\usepackage{listings}
\usepackage{amsfonts,amssymb}
\usepackage{bm}
\usepackage{graphicx}
\usepackage{lipsum}

\usepackage[accsupp]{axessibility}
\usepackage{colortbl}

\newlength\savedwidth

\usepackage{verbatim}
\usepackage{tikz}
\usepackage{textpos}

\definecolor{resnet}{rgb}{0.9607843137254902, 0.8705882352941177, 0.7019607843137254}
\definecolor{regnety}{rgb}{0.9764705882352941 , 0.792156862745098 , 0.1411764705882353}
\definecolor{vit}{rgb}{0.9019607843137255 , 0.403921568627451 , 0.403921568627451}
\definecolor{deit}{rgb}{1.0 , 0.7137254901960784 , 0.7568627450980392}
\definecolor{unet}{rgb}{0.7686274509803922 , 0.27058823529411763 , 0.4117647058823529}
\definecolor{mlpmixer}{rgb}{0.0 , 0.5803921568627451 , 0.19607843137254902}
\definecolor{resmlp}{rgb}{0.25098039215686274 , 0.8784313725490196 , 0.8156862745098039}
\definecolor{mamba}{rgb}{0.2196078431372549 , 0.403921568627451 , 0.8392156862745098}
\definecolor{gmlp}{rgb}{0.0 , 0.7215686274509804 , 0.5803921568627451}
\definecolor{dit}{rgb}{0.8666666666666667 , 0.6274509803921569 , 0.8666666666666667}
\definecolor{resnetimproved}{rgb}{1.0 , 0.54902 , 0.0}

\newcommand{\unetdot}{\raisebox{0.5pt}{\tikz\fill[unet] (0,0) (0ex,1ex)--(-0.866ex,-0.5ex)--(0.866ex, -0.5ex)--cycle;}}
\newcommand{\mlpmixerdot}{\raisebox{0.5pt}{\tikz\fill[mlpmixer] (0,0) (1ex,0ex)--(-0.5ex,0.866ex)--(-0.5ex, -0.866ex)--cycle;}}

\newcommand{\mamba}{\raisebox{0.5pt}{\tikz\fill[mamba] (0,0) (1ex,0ex)--(-0.5ex,0.866ex)--(-0.5ex, -0.866ex)--cycle;}}

\newcommand{\dit}{\raisebox{0.5pt}{\tikz\fill[dit] (0,0) circle (.8ex);}}

\def\eg{\emph{e.g}\onedot} 
\def\ie{\emph{i.e}\onedot} 
 
\def\etc{\emph{etc}\onedot}

\newcommand{\gr}{\rowcolor[gray]{.95}}

\definecolor{iccvblue}{rgb}{0.21,0.49,0.74}
\usepackage[pagebackref,breaklinks,colorlinks,allcolors=iccvblue]{hyperref}

\def\paperID{8218} % *** Enter the Paper ID here
\def\confName{ICCV}
\def\confYear{2025}

\title{LiT: Delving into a Simple Linear Diffusion Transformer for Image Generation}

\author{
Jiahao Wang\textsuperscript{1,2}\footnotemark[2] 
\quad Ning Kang\textsuperscript{3}
\quad Lewei Yao\textsuperscript{3}
\quad Mengzhao Chen\textsuperscript{1}
\quad Chengyue Wu\textsuperscript{1}
\\
\quad Songyang Zhang\textsuperscript{2}
\quad Shuchen Xue\textsuperscript{4}\thanks{Work done when Shuchen Xue interns at Huawei.}
\quad Yong Liu\textsuperscript{5}
\quad Taiqiang Wu\textsuperscript{1}
\quad Xihui Liu\textsuperscript{1}
\\
\quad Kaipeng Zhang\textsuperscript{2} 
\quad Shifeng Zhang\textsuperscript{3} 
\quad Wenqi Shao\textsuperscript{2}\thanks{Correspondence to: Jiahao Wang (wang-jh19@tsinghua.org.cn), Wenqi Shao (shaowenqi@pjlab.org.cn) and Zhenguo Li (li.zhenguo@huawei.com).}
\quad Zhenguo Li\textsuperscript{3}\footnotemark[2] 
\quad Ping Luo\textsuperscript{1}
\\
\textsuperscript{1}{HKU}  \quad
\textsuperscript{2}{Shanghai AI Lab}  \quad 
\textsuperscript{3}{Huawei Noah's Ark Lab}  \quad 
\textsuperscript{4}{UCAS}  \quad 
\textsuperscript{5}{THUsz} \\
\tt\small wang-jh19@tsinghua.org.cn \quad shaowenqi@pjlab.org.cn \quad li.zhenguo@huawei.com \\
}

\begin{document}
\maketitle
\begin{abstract}

In this paper, we investigate how to convert a pre-trained Diffusion Transformer (DiT) into a linear DiT, as its simplicity, parallelism, and efficiency for image generation. 
Through detailed exploration, we offer a suite of ready-to-use solutions, ranging from linear attention design to optimization strategies. 
Our core contributions include 5 practical guidelines:
1) Applying depth-wise convolution within simple linear attention is sufficient for image generation.
2) Using fewer heads in linear attention provides a free-lunch performance boost without increasing latency.
3) Inheriting weights from a fully converged, pre-trained DiT.  
4) Loading all parameters except those related to linear attention. 
5) Hybrid knowledge distillation: using a pre-trained teacher DiT to help the training of the student linear DiT, supervising not only the predicted noise but also the variance of the reverse diffusion process. 
These guidelines lead to our proposed \underline{L}inear D\underline{i}ffusion \underline{T}ransformer (LiT), which serves as a safe and efficient alternative baseline for DiT with pure linear attention.
In class-conditional 256$\times$256 and 512$\times$512 ImageNet generation, LiT can be quickly adapted from DiT using only $20\%$ and $33\%$ of DiT’s training steps, respectively, while achieving comparable performance. LiT also rivals methods based on Mamba or Gated Linear Attention. Moreover, the same guidelines generalize to text-to-image generation: LiT can be swiftly converted from PixArt-$\Sigma$ to generate high-quality images, maintaining comparable GenEval scores. 
% Additionally, LiT supports offline deployment on a laptop, enabling 1K resolution photorealistic image generation. 

% allows for the rapid synthesis of up to 1K resolution photorealistic images. 

% Experiments show that in class-conditional 256$\times$256 and 512$\times$512 ImageNet benchmark LiT achieves highly competitive FID while reducing training steps by $80\%$ and $77\%$ compared to DiT. 
% LiT also rivals methods based on Mamba or Gated Linear Attention. 
% Besides, for text-to-image generation, LiT allows for the rapid synthesis of up to 1K resolution photorealistic images. 
% Project page: \url{https://techmonsterwang.github.io/LiT/}.

% \small\blfootnote{\noindent$^{\ast}$ Corresponding author.}

\end{abstract}    
\section{Introduction}
\label{sec:intro}

% Diffusion models~\cite{ho2020denoising, sohl2015deep} based on Transformers~\cite{chen2024pixart3, esser2024scaling, chen2024pixart1}, demonstrate potential commercial value in the field of image generation, but suffer from latency and GPU memory usage in high-resolution scenarios. This is due to its main self-attention modules exhibits a quadratic complexity with respect to sequence length, which makes it less feasible for resource-limited edge devices, as illustrated in Fig.~\ref{fig:res_cos_sim} and Fig.~\ref{fig:latency} (Sec.~\ref{sec:appendix.2} of the appendix).

Diffusion models~\cite{ho2020denoising, sohl2015deep} based on Transformers~\cite{chen2024pixart3, esser2024scaling, chen2024pixart1} show strong commercial potential in image generation but face latency and GPU memory challenges in high-resolution tasks. Their main self-attention modules exhibit quadratic complexity with respect to sequence length, making deployment on resource-limited edge devices less feasible, as illustrated in Fig.~\ref{fig:res_cos_sim} and Fig.~1 in Appendix Sec.~B, highlighting the need for sub-quadratic attention modules. 

\begin{figure}[t]
	\centering
	\includegraphics[width=0.65\linewidth]{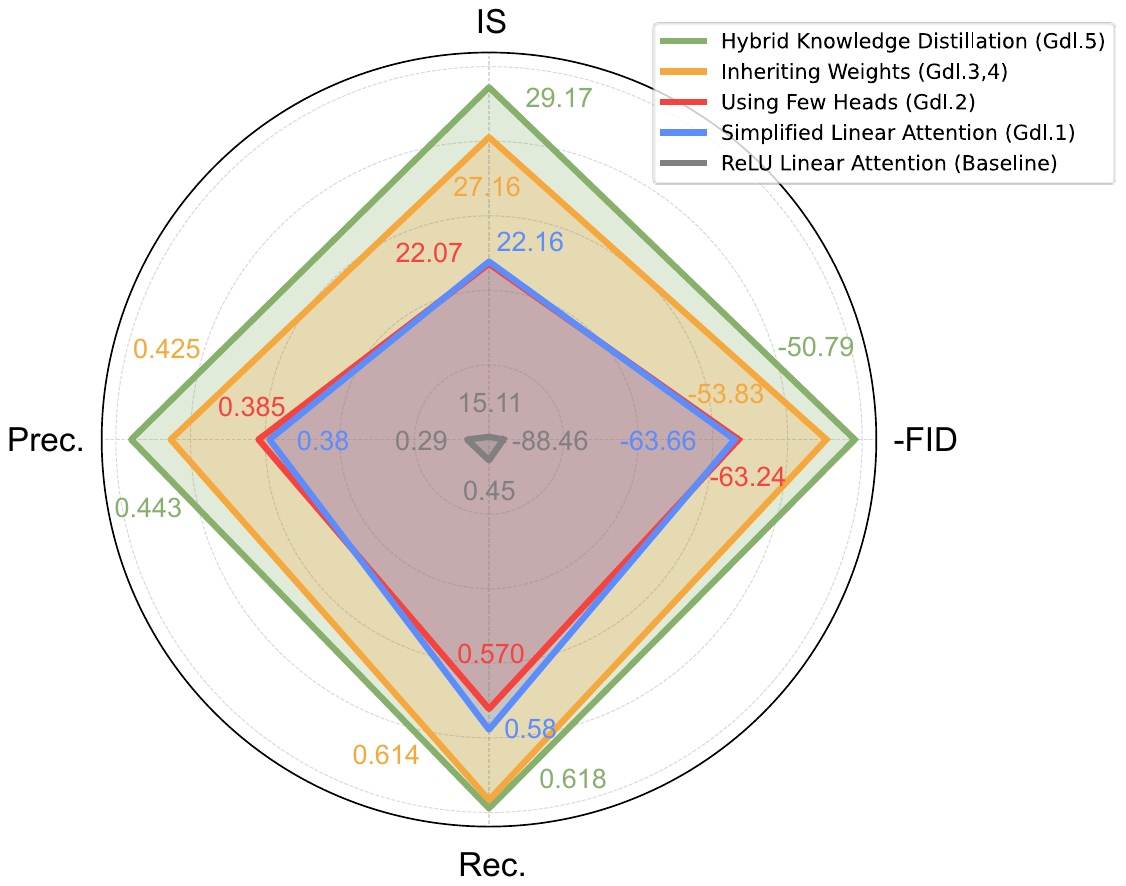}
    \vspace{-0.1in}
	\caption{\textbf{Roadmap for improving LiT}. We present five practical guidelines, covering both architectural designs and training strategies, to enhance the generative performance of linear attention.}
	\label{fig:radar_new}
    \vspace{-0.14in}
\end{figure}

\begin{figure*}[t]
	\centering
	\includegraphics[width=1.0\linewidth]{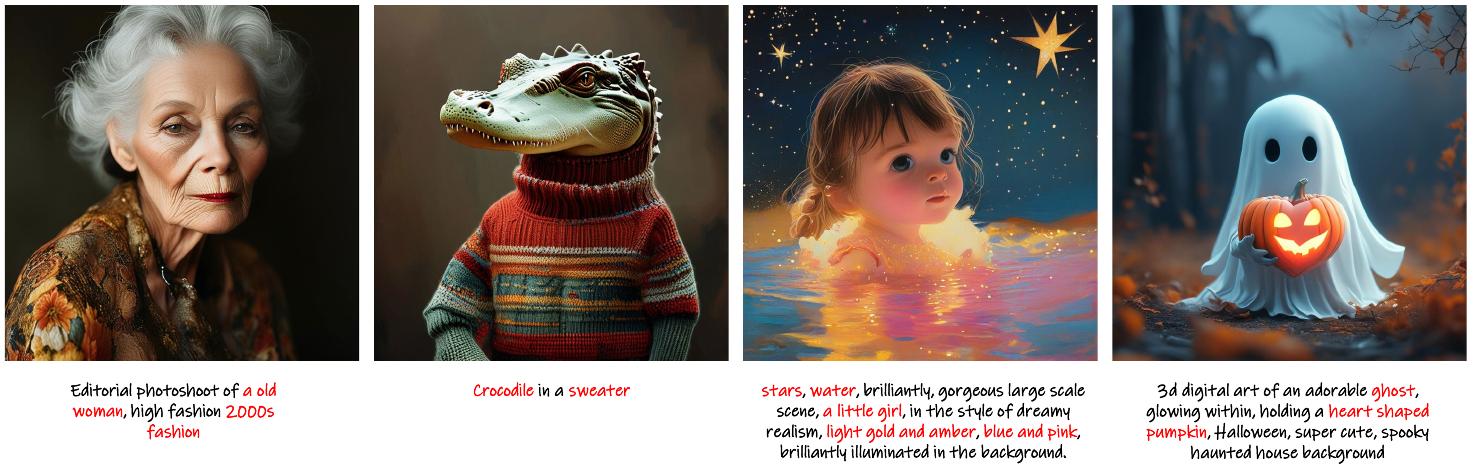}
    \vspace{-0.2in}
	\caption{\textbf{1K Generated samples of LiT following user instructions.} Converted from PixArt-$\Sigma$~\cite{chen2024pixart3}, LiT adopts the same macro- and micro-level architecture, maintaining alignment with the PixArt-$\Sigma$ framework while elegantly replacing all self-attention with efficient linear attention. While being simple and efficient, LiT can generate exceptional high-resolution images following user instructions.}
	\label{fig:t2i-outputs-1k}
    \vspace{-0.2in}
\end{figure*}

Linear attention can significantly accelerate image generation compared to self-attention at the same resolution. For example, at 2048px, it is nearly 9$\times$ faster (Fig.~\ref{fig:res_cos_sim}-(a), (b)). Additionally, it substantially reduces GPU memory usage for high-resolution image generation. Replacing self-attention with linear attention in DiT-S/2 reduces memory consumption from $~\sim$14GB to 4GB at 2048px. In fact, linear attention has been extensively validated in visual recognition tasks (\eg, image classification and segmentation)~\cite{cai2023efficientvit, han2023flatten, bolya2022hydra}. In these methods, the linear attention is expected to generate sharp distributions that effectively capture semantic information~\cite{cai2023efficientvit, han2023flatten}. However, in the context of denoising diffusion models, linear attention may require enhanced locality to better model the distribution of random noise. Consequently, exploring the use of linear attention in image generation remains a promising research direction.

Several efforts have explored the use of efficient attention for image generation.
For instance, 
DiG~\cite{zhu2024dig} and DiM~\cite{teng2024dim} reference Gated Linear Attention (GLA)~\cite{yang2023gated} and State Space Models (SSM)~\cite{gu2023modeling, gu2023mamba} to implement diffusion models, respectively. 
% In fact, the original GLA and SSM are designed for 1-D language sequences, requiring specific adaptations when applied to 2-D images~\cite{zhu2024dig, teng2024dim}. 
SANA~\cite{xie2024sana} takes a deep compression autoencoder~\cite{chen2024deep}, and replace the original MLP-FFN with Mix-FFN~\cite{cai2023efficientvit}. 
Attention Mediators~\cite{pu2024efficient} applies a set of additional tokens in linear attention. 
CLEAR~\cite{liu2024clear} adopts a circular window within the attention.

\begin{figure*}[t]
	\centering
	\includegraphics[width=1.0\linewidth]{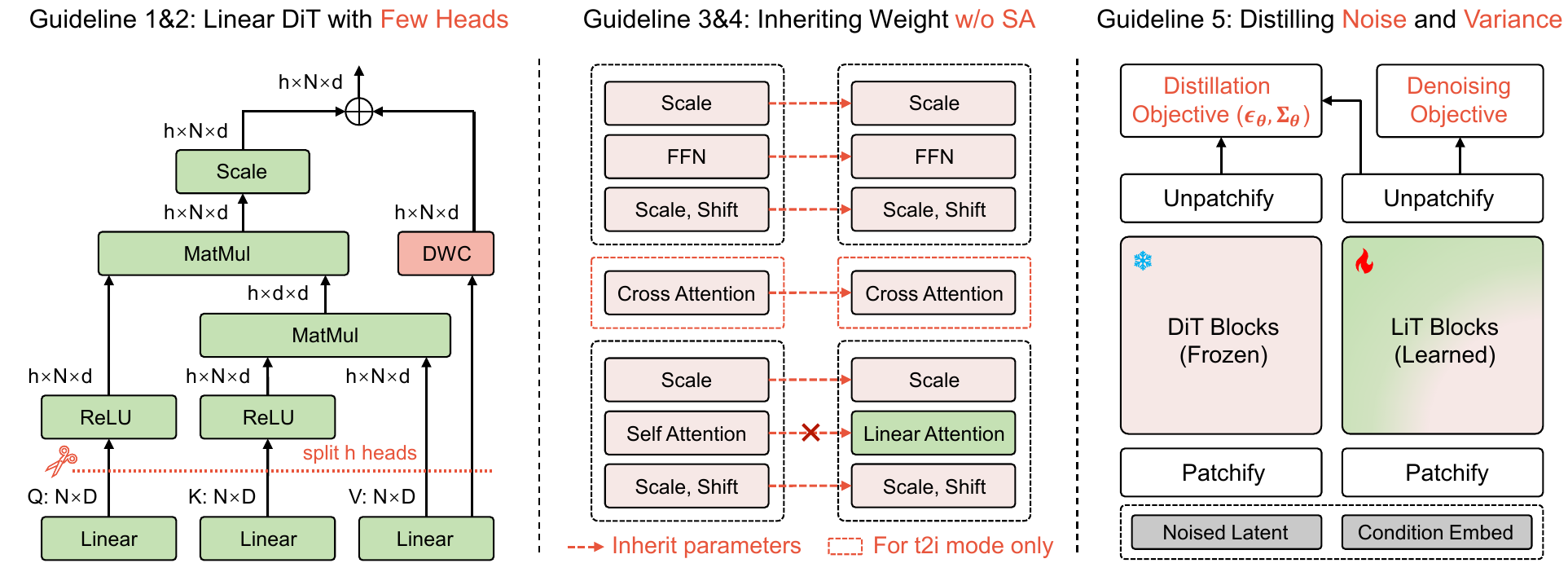}
    \vspace{-0.2in}
	\caption{\textbf{Overall training procedure of LiT.} Converted from DiT~\cite{peebles2023scalable} (for class-conditioned image generation) or PixArt-$\Sigma$~\cite{chen2024pixart3} (for text-to-image generation), LiT keeps their macro/micro-level design but safely and efficiently using pure simple linear attention. LiT is built up with (1) building a strong linear DiT baseline with few heads, (2) inheriting weights from a DiT teacher and (3) distilling useful knowledge (predicted noise and the variances of the reverse diffusion process) from the teacher model.}
	\label{fig:architecture}
    \vspace{-0.2in}
\end{figure*}

In this paper, we systematically investigate the problem of converting a pre-trained DiT to a linear DiT for image generation while minimizing performance loss. We decompose this problem into two parts: first, \textit{what type of linear attention architecture is suitable for image generation?} Second, \textit{what training strategies are effective for optimizing a linear DiT?}
To address these two questions, we conducted extensive experiments on Linear DiT from both \textit{architecture design} and \textit{training strategy} perspectives. Based on the results, our findings can be summarized into \textbf{five practical guidelines} for Linear DiT design and optimization:
\textbf{1}) simply adding a depth-wise convolution in linear attention is sufficient for image generation; 
\textbf{2}) a free lunch: few (\eg, 2) heads in linear attention improves performance without increasing latency; 
\textbf{3}) initialize the weights of the linear DiT from a well-trained DiT; 
\textbf{4}) load all weights except for the linear attention to achieve cost-effective training; 
\textbf{5}) hybrid distillation objective—distilling not only the predicted noise but also variances of the reverse diffusion process—to facilitate the student linear DiT. Following such guidelines (Fig.~\ref{fig:architecture}), the result diffusion model, dubbed \textit{\underline{L}inear D\underline{i}ffusion \underline{T}ransformer (LiT)}, can adopt pure \textit{simple} attention design, achieving \textit{linear} computational complexity while retaining the \textit{capability} of full attention. 
Meanwhile, LiT is trained in a highly cost-effective manner as it is derived from a pre-trained DiT, leveraging existing knowledge and reducing the need for extensive training from scratch.

LiT is an affirmative answer to how to efficiently and safely obtain a linear DiT from a pretrained DiT. To validate its effectiveness, we conduct experiments on representative diffusion Transformers, \ie, DiT (for ImageNet~\cite{deng2009imagenet} class-conditional image generation) and PixArt-$\Sigma$~\cite{chen2024pixart3} (for text-to-image generation).
In ImageNet~\cite{deng2009imagenet} 256$\times$256 benchmark, LiT-S/B/L (patch size of 2) with only 100K training steps outperforms corresponding 400K steps pre-trained DiT-S/B/L in terms of Fréchet Inception Distance (FID)~\cite{heusel2017gans}.
For large-scale model on the ImageNet 256$\times$256 and 512$\times$512 benchmark, LiT-XL/2 competes comparably to DiT-XL/2 (2.32 vs. 2.27, 3.69 vs. 3.04 in FID) with only $20\%$ and $\sim 23\%$ of training steps (respectively for 256$\times$256 and 512$\times$512 settings). In text-to-image generation, converted from PixArt-$\Sigma$, LiT-0.6B shares the same macro/micro-level architecture, ensuring alignment with the PixArt-$\Sigma$ framework while elegantly replacing all self-attention with efficient linear attention. LiT-0.6B can stably generate highly photorealistic images (Fig.~\ref{fig:t2i-outputs-1k}). To verify the effectiveness of linear attention, we deploy LiT-0.6B offline on a Windows 11 laptop (Fig.~\ref{fig:laptop_dep}), enabling 1K-resolution image synthesis based on human instructions.

% \begin{itemize}[leftmargin=*,topsep=0pt,itemsep=0pt,noitemsep]
% \item We introduce a \textit{linear diffusion Transformer}, LiT, which optimizes the latency-capacity trade-off by using a \textit{few heads} (\eg, 2) in linear attention.
% \item We initialize LiT with weight inheritance to ensure a \textit{cost-effective} training process, while achieving linear computational complexity that facilitates high-resolution generation scenarios during inference.
% \item We propose a \textit{hybrid} distillation approach, thereby successfully apply LiT to class-conditional image generation and text-to-image generation. 
% \item The pretrained LiT-0.6B model can be deployed on a Windows 11 laptop to generate high-resolution, photo-realistic images offline based on user prompts.
% \end{itemize}

% 我们提出了一个线性扩散Transformer，PixArt,并通过使用很少的heads优化了线性注意力的capacity和时延。
% 我们借助权重继承来优化PixArt，使其训练过程cost-effective，同时推理时具备对高分辨率使用场景友好的线性计算复杂度。
% 我们提出一种用于扩散模型的混合知识蒸馏训练目标，并成功将PixArt应用于图像生成和文生图任务。我们以PixArt为例为社区总结了训练线性扩散模型的5 guidelines。

% \input{sec/2_related_work}
\section{Preliminaries}
\label{sec:preliminaries}

% \subsection{Preliminaries}
% \label{sec:preliminaries.1}
\paragraph{Diffusion models.} Suppose we have a clean image $\mathbf{x}_{0}$, and progressively add noise at each step $t$ to obtain $\mathbf{x}_{T}, ~t\in\left[ 1,T \right]$. The forward (diffusion) process is defined as: $q(\mathbf{x}_t \vert \mathbf{x}_{t-1}) = \mathcal{N}(\mathbf{x}_t; \sqrt{1 - \beta_t} \mathbf{x}_{t-1}, \beta_t\mathbf{I})$, where $\alpha_t=1-\beta_t$ and $\bar{\alpha}_t = \prod_{i=1}^t \alpha_i$. Diffusion models~\cite{ho2020denoising, sohl2015deep, luo2022understanding, weng2021diffusion} learn a network $p_{\theta}$ to reverse the diffusion process: $p_\theta(\mathbf{x}_{t-1} \vert \mathbf{x}_t) = \mathcal{N}(\mathbf{x}_{t-1}; \boldsymbol{\mu}_\theta(\mathbf{x}_t, t), \boldsymbol{\Sigma}_\theta(\mathbf{x}_t, t))$, where $\boldsymbol{\mu}_\theta(\mathbf{x}_t, t)$ and $\boldsymbol{\Sigma}_\theta(\mathbf{x}_t, t)$ denote the mean and variance of the reverse process. In DDPM~\cite{ho2020denoising}, a simple training objective is demonstrated effective in training diffusion models:
$L_t^\text{simple}
= \mathbb{E}_{t, \mathbf{x}_0, \boldsymbol{\epsilon}_t} \Big[\|\boldsymbol{\epsilon}_t - \boldsymbol{\epsilon}_\theta(\sqrt{\bar{\alpha}_t}\mathbf{x}_0 + \sqrt{1 - \bar{\alpha}_t}\boldsymbol{\epsilon}_t, t)\|^2 \Big]$, 
where $\boldsymbol{\epsilon}_t$ represents noise of the $t$ step, expected to be approximated by the output of a denoising network $\boldsymbol{\epsilon}_\theta$. $\boldsymbol{\epsilon}_\theta$ can be realized using Transformers~\cite{peebles2023scalable, ma2024sit, chen2024pixart1, chen2024pixart3} with time-intensive self-attention, calling for cheap alternatives.

\begin{table}
    \begin{center}
        \small
        \begin{tabular}{lcccc}
            \hline
            DiT & Attention  & FID-50K ($\downarrow$) & IS ($\uparrow$)	\\
            \hline
            S/2    & Softmax & 68.40 & -  \\
            
            S/2    & ReLU Linear Baseline & 88.46 & 15.11  \\

            \gr
            S/2    & + Depth-wise Conv. (ReLU) & 63.66 & 22.16  \\
            
            S/2    & + Focused Linear (ReLU) & \textbf{63.05} & \textbf{22.49}  \\

            S/2    & + Focused Linear (GELU) & 70.83 & 19.41  \\

            \hline
            B/2    & softmax & 43.47 & -  \\
            
            B/2    & ReLU Linear Baseline & 56.92 & 25.80  \\

            \gr
            B/2    & + Depth-wise Conv. (ReLU) & 42.11 & 34.60  \\
            
            B/2    & + Focused Linear (ReLU) & \textbf{40.58} & \textbf{35.98}  \\

            B/2    & + Focused Linear (GELU) & 58.86 & 24.23  \\
            
            \hline
        \end{tabular}
    \end{center}
    \vspace{-0.2in}
    \caption{\textbf{Linear attention ablation on ImageNet 256$\times$256}. All models are DiT trained for 400K steps. We report FID-50K (\textit{without} classifier-free guidance) and Inception Score metrics.}
    \label{table-4-1}
    \vspace{-0.2in}
\end{table}

\vspace{-10pt}

\paragraph{Linear attention.} For a standard softmax attention with $h$ heads, sequence length $N$, and hidden dimension $D=h\cdot d$, denoting $\mathbf{Q}, \mathbf{K}, \mathbf{V} \in\mathbb{R}^{N\times D}$ as query, key and value, its output $\mathbf{O} \in\mathbb{R}^{N\times D}$ can be expressed as:
\begin{equation} 
\small
\mathbf{O}_i=\sum_{j=1}^{N}\ \frac{{\rm Sim}{\left(\mathbf{Q}_i,\mathbf{K}_j\right)}}{\sum_{j=1}^{N}\ {\rm Sim}{\left(\mathbf{Q}_i,\mathbf{K}_j\right)}}\mathbf{V}_j,
\label{equ:preliminaries_1}
\end{equation}
where ${\rm Sim}\left(\mathbf{Q},\mathbf{K}\right)={\rm exp}({\mathbf{Q}\mathbf{K}^\top}/{\sqrt d})$ is a similarity measurement function. In linear attention, the aforementioned function is modified into a simplified form with a nonlinear kernel $\boldsymbol{\phi}(\cdot)$, \ie, ${\rm Sim}\left(\mathbf{Q},\mathbf{K}\right)=\boldsymbol{\phi}(\mathbf{Q})\boldsymbol{\phi}(\mathbf{K})^\top$. Thus, Eq.~\ref{equ:preliminaries_1} can be rewritten as Eq.~\ref{equ:preliminaries_2} and simplified to Eq.~\ref{equ:preliminaries_3} using the associative property of multiplication.
\begin{equation}
\small
\mathbf{O}_i=\sum_{j=1}^{N}\ \frac{\boldsymbol{\phi}\left(\mathbf{Q}_i\right)\boldsymbol{\phi}\left(\mathbf{K}_j\right)^\top}{\sum_{j=1}^{N}{\boldsymbol{\phi}\left(\mathbf{Q}_i\right)\boldsymbol{\phi}\left(\mathbf{K}_j\right)^\top\ }}\mathbf{V}_j,
\label{equ:preliminaries_2}
\end{equation}
\begin{equation}
\small
\mathbf{O}_i=\frac{\boldsymbol{\phi}(\mathbf{Q}_i)\left(\sum_{j=1}^{N}{\boldsymbol{\phi}{(\mathbf{K}_j)}^\top\mathbf{V}_j\ }\right)}{\boldsymbol{\phi}(\mathbf{Q}_i)\left(\ \sum_{j=1}^{N}{\boldsymbol{\phi}{(\mathbf{K}_j)}^\top\ }\right)},
\label{equ:preliminaries_3}
\end{equation}
The computational complexities of softmax attention and linear attention are $\mathcal{O}(N^2D)$ and $\mathcal{O}(ND^2/h)$, respectively. 
% At 1K resolution, using a latent diffusion Transformer~\cite{rombach2022high} with an 8$\times$ downsampling VAE~\cite{kingma2013auto} encoder (as in DiT-XL/2~\cite{peebles2023scalable}), sequence length is 4096. Accordingly, we have $\frac{N^2D}{ND^2/h}\approx56.9$. Detailed presentation is provided in Sec.~\ref{sec:method.2}. 

Despite its efficiency, the suitability of linear attention for image generation remains under-explored. Most existing works, such as SANA~\cite{xie2024sana}, Mediators~\cite{pu2024efficient}, and CLEAR~\cite{liu2024clear}, introduce carefully designed modifications to DiT, compromising its simplicity. Motivated by this, we explore how to safely and efficiently convert a pretrained DiT into a linear DiT without relying on complex architectures or sacrificing performance.

\section{Exploring LiT: A Roadmap}
\label{sec:method}

We explore the architectural design and optimization strategies of our LiT based on DiT~\cite{peebles2023scalable} on class-conditional ImageNet~\cite{deng2009imagenet} 256$\times$256 benchmark, as it is relatively simple, thereby highlighting the ablation effect of attention itself. We train models for 400K iterations using a batch size of 256, and report the FID-50K~\cite{heusel2017gans}, Inception Score~\cite{salimans2016improved} and Precision/Recall~\cite{kynkaanniemi2019improved} metrics. The radar chart of the experimental results is shown in Fig.~\ref{fig:radar_new}. Detailed results can be found in Sec.~E of the Appendix. 

\subsection{Architectural Guidelines for Linear DiT}
\label{sec:method.1}

\begin{figure}[t]
\vspace{-0.0em}
\centering
\begin{tabular}{cc}
\includegraphics[width=0.456\linewidth]{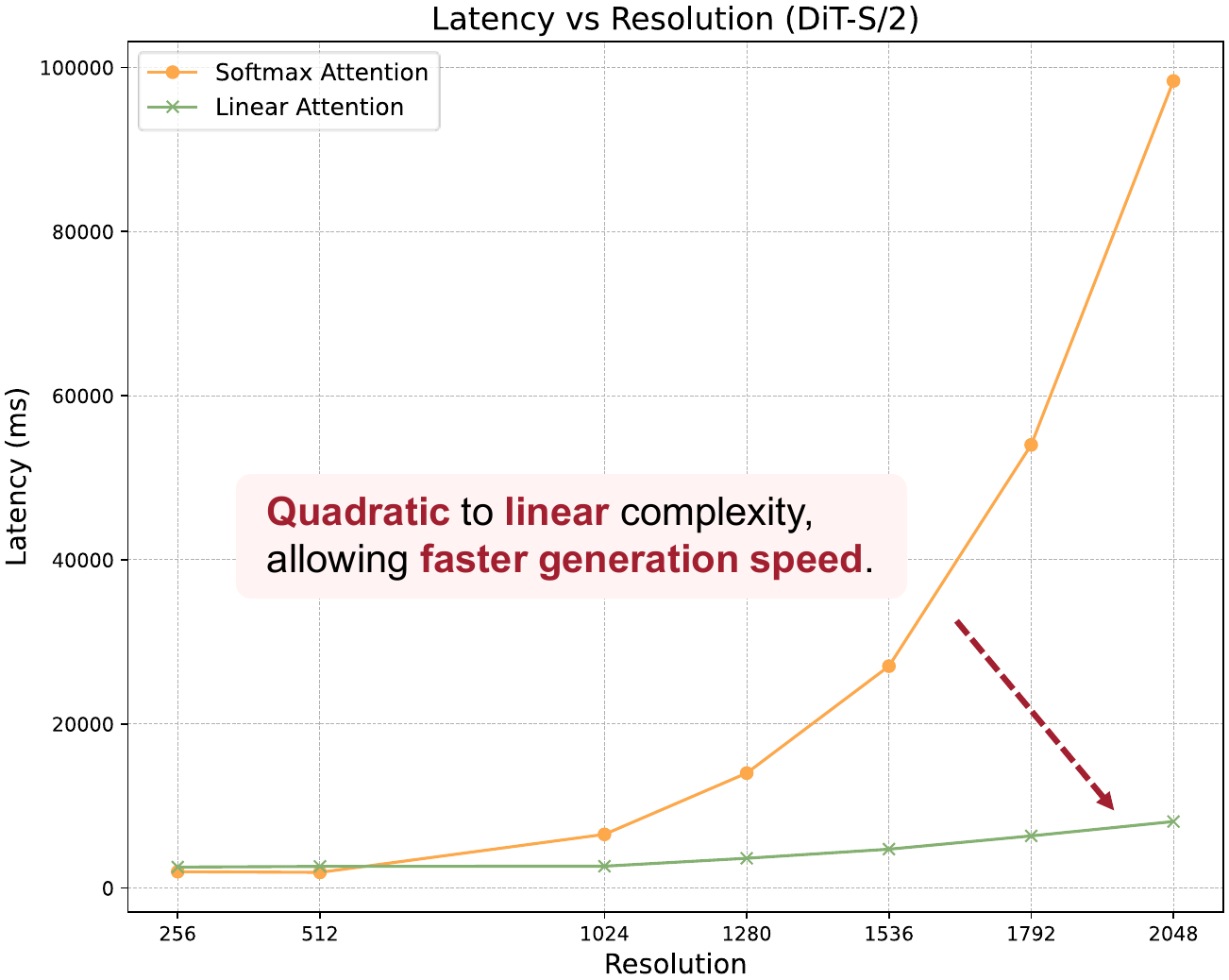} &
\includegraphics[width=0.454\linewidth]{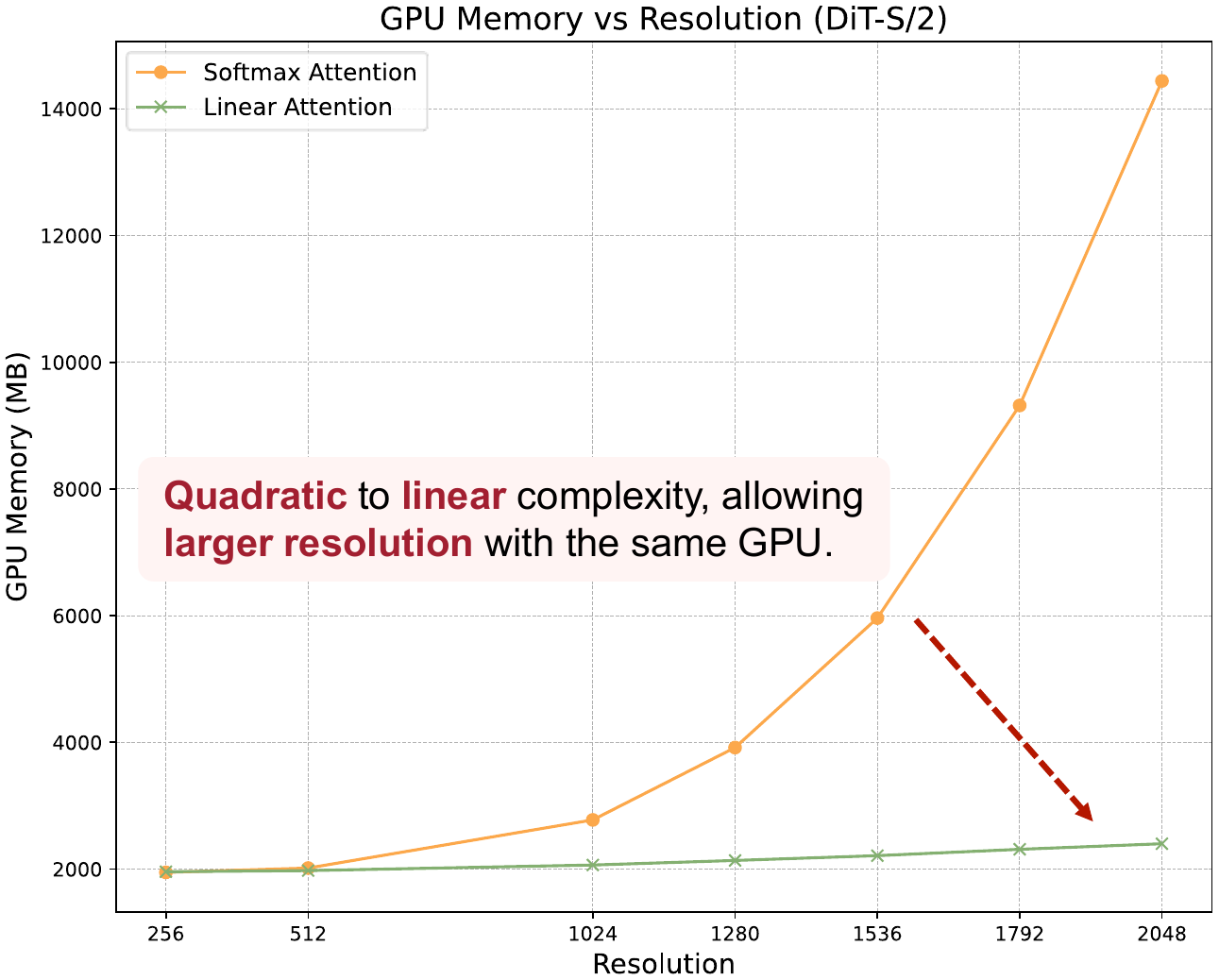}  \\
~~~~(a) Latency & ~~~ (b) GPU Memory \\
\end{tabular}
\vspace{-0.1in}
\caption{\textbf{Efficiency of linear attention.} Comparison of (a) module latency and (b) GPU memory between {\color{softmax}{DiT-S/2}} and {\color{linear}{linear DiT-S/2}}  in synthesizing images of different resolutions. 
We test on one NVIDIA A100 GPU with a batch size of 1, and do not use any extra acceleration techniques for attention (\eg, flash attention~\cite{dao2022flashattention}). 
}
\label{fig:res_cos_sim}
\vspace{-0.1in}
\end{figure}

\begin{table}
    \begin{center}
        \small
        \scalebox{0.977}{
        \begin{tabular}{lcccccc}
            \hline
            DiT & Head  & FID-50K ($\downarrow$) & IS ($\uparrow$) & Prec. ($\uparrow$) & Rec. ($\uparrow$)	\\
            \hline
            S/2    & 1 & 64.42 & 21.54 & 0.380 & 0.574  \\

            \gr
            S/2    & 2 & 63.24 & 22.07 & 0.385 & 0.570  \\

            S/2    & 3 & \textbf{63.21} & \textbf{22.08} & \textbf{0.386} & \textbf{0.583}  \\

            S/2    & \underline{6} & 63.66 & 22.16 & 0.383 & 0.580  \\

            S/2    & 48 & 78.76 & 17.46 & 0.322 & 0.482  \\

            S/2    & 96 & 116.00 & 11.49 & 0.224 & 0.261  \\

            \hline
            B/2    & 1 & 41.77 & 34.78 & 0.487 & 0.631  \\
 
            B/2    & 2 & 41.39 & 35.59 & 0.494 & 0.631  \\
            
            \gr
            B/2    & 3 & \textbf{40.86} & \textbf{35.79} & \textbf{0.497} & \textbf{0.629} \\

            B/2    & \underline{12} & 42.11 & 34.60 & 0.484 & 0.631  \\

            B/2    & 96 & 68.30 & 20.45 & 0.375 & 0.531  \\

            B/2    & 192 & 112.39 & 12.07 & 0.240 & 0.282  \\
            \hline
            L/2    & 1 & 24.46 & 57.36 & 0.600 & 0.637  \\

            L/2    & 2 & 24.37 & 57.02 & 0.599 & 0.622 \\

            \gr
            L/2    & 4 & \textbf{24.04} & \textbf{59.02} & \textbf{0.597} & \textbf{0.636}  \\

            L/2    & \underline{16} & 25.25 & 54.67 & 0.587 & 0.632  \\

            \hline
            XL/2    & 1 & 21.13 & 65.06 & 0.619 & 0.632  \\

            \gr
            XL/2    & 2 & \textbf{20.66} & \textbf{65.39} & \textbf{0.624} & \textbf{0.636}  \\

            \gr
            XL/2    & 4 & 20.82 & 65.52 & 0.619 & 0.632  \\

            XL/2    & \underline{16} & 21.69 & 63.06 & 0.617 & 0.628  \\

            \hline
        \end{tabular}}
    \end{center}
    \vspace{-0.2in}
    \caption{\textbf{Head number ablation of linear attention on ImageNet 256$\times$256}. We report FID-50K (\textit{without} classifier-free guidance), Inception Score, Precision/Recall metrics. \underline{DiTs setting}.}
    \label{table-4-2}
    \vspace{-0.2in}
\end{table}

\paragraph{Guideline 1: Simply adding a depth-wise convolution in linear attention is sufficient for DiT-based image generation.} 
In high-level visual recognition tasks (\eg, image classification, instance segmentation, \etc), 
linear attention has been reported to produce a ``smooth" attention distribution compared to the ``sharp" distribution of softmax attention~\cite{cai2023efficientvit, han2023flatten}. 
To address the issue, Flatten Transformer~\cite{han2023flatten} introduces a focused function applied after the ReLU activation as a remedy. 
Meanwhile, linear attention has been shown to exhibit limited feature diversity due to its low-rank attention matrix. Accordingly, only a single depth-wise convolution layer, applied on the value matrix, can raise the upper bound of the equivalent rank, effectively mitigating the low-rank issue~\cite{han2023flatten}. 
Based on these observations, it is reasonable to validate similar conclusions in image generation. 
Compared to the full attention baseline, LiT-S/2 and B/2 with ReLU linear attention exhibit an unacceptable performance drop (Tab.~\ref{table-4-1}), with FID increasing by approximately 20/13 for the S/2 and B/2 models, respectively. 
We then explore the following approaches: (1) adding a depth-wise convolution (DWC)~\cite{howard2017mobilenets, chollet2017xception} on the value matrix\footnote{In this setup, we optimize the linear attention implementation (\eg, by replacing \texttt{einsum} with the ``@" operator for matrix operations and removing the softplus function). We also test a version without these optimizations for comparison, which yielded results of 63.13 (S/2) and 42.13 (B/2), showing only slight differences from the results in Tab.~\ref{table-4-1}: 63.66 (S/2) and 42.11 (B/2). Therefore, we retain the optimizations.}; (2) adding focused function~\cite{han2023flatten}; (3) replacing the ReLU kernel with GELU~\cite{hendrycks2016gaussian} function, commonly used in Transformers~\cite{vaswani2017attention}. Each of these choices maintains linear complexity, preserving LiT’s advantage in computational efficiency. We use a relatively large convolution kernel (\ie, 5) to ensure a sufficiently large receptive field when predicting noise.

As shown in Tab.~\ref{table-4-1}, incorporating DWC alone enhances generation quality beyond softmax attention, achieving FID scores of 63.66 and 42.11 for S/2 and B/2, respectively.
We attribute this improvement to the model's tendency to rely on information from neighboring pixels when predicting noise for a given pixel, necessitating the assistance of convolution.
Meanwhile, we identify that the focused function has limited effectiveness, which we attribute to its design motivation to help linear attention focus on specific regions. 
This feature may suit classification models, but may not be necessary for noise prediction.

\noindent \textbf{Remark 1.} We note that our findings on linear attention in visual generation align with those of the SLAB~\cite{guo2024slab} in visual recognition, where simple DWC leads to improvement. We attribute this to linear attention's lack of locality modeling capacity compared to full attention, which DWC serves as a remedy. We will adopt this design afterwards.

\begin{table}
    \begin{center}
        \small
        \begin{tabular}{lccccc}
            \hline
            Load & Iter. & FFN & Modu. & Attention  & FID-50K ($\downarrow$) \\
            \hline
            
            \gr
            model  & 400K & \ding{51} & \ding{51} & \ding{55} & 56.07 \\

            ema  & 400K & \ding{51} & \ding{51} & \ding{55} & 56.07 \\

            \hline

            model  & 200K & \ding{51} & \ding{51} & \ding{55} & 57.84 \\

            model  & 300K & \ding{51} & \ding{51} & \ding{55} & 56.95 \\

            model  & 400K & \ding{51} & \ding{51} & \ding{55} & 56.07 \\

            model  & 600K & \ding{51} & \ding{51} & \ding{55} & 54.80 \\

            \gr
            model  & 800K & \ding{51} & \ding{51} & \ding{55} & \textbf{53.83} \\

            \hline

            model  & 600K & \ding{51} & \ding{51} & Q, K, V & 55.29 \\

            model  & 600K & \ding{51} & \ding{51} & K, V & 55.07 \\

            model  & 600K & \ding{51} & \ding{51} & V & 54.93 \\

            model  & 600K & \ding{51} & \ding{51} & Q & 54.82 \\

            model  & 600K & \ding{51} & \ding{51} & O & 54.84 \\
            
            \hline
        \end{tabular}
    \end{center}
    \vspace{-0.2in}
    \caption{\textbf{Weight inheritance ablation on ImageNet 256$\times$256}. Baseline linear DiT-S/2 (using 2 heads) are trained for 400K steps.}
    \label{table-4-3}
    \vspace{-0.2in}
\end{table}

\subsection{Linear Attention with Few Heads}
\label{sec:method.2}

% \begin{figure*}[t]
% \vspace{-0.0em}
% \centering
% \begin{tabular}{ccc}
% \includegraphics[width=0.44\linewidth]{figures/head_small_base.pdf} &~
% \includegraphics[width=0.444\linewidth]{figures/head_large_xlarge.pdf} \\
% (a) Results of small (S) and base (B) model & ~~~ (b) Results of large (L) and Xlarge (XL) model  \\
% \end{tabular}
% \vspace{-0.2em}
% \caption{\textbf{Free lunch in linear attention.} Comparison of {\color{latency}{latency}} and theoretical {\color{gmacs}{GMACs}} for linear attention with different number of heads. We test the latency to generate $256\times256$ resolution images using one NVIDIA A100 GPU with a batch size of 8. Results of (a) S/2/B/2 and (b) L/2/XL/2 models were averaged over 30 times. As the number of heads decreases, GMACs consistently increase, but no ascending trend in latency has been observed. Based on the observation, we use few heads (\eg, 2) to get the free lunch in linear attention. 
% }
% \label{fig:free_lunch}
% \vspace{-0.3em}
% \end{figure*}

\begin{figure}[t]
\vspace{-0.0em}
\centering
\begin{tabular}{cc}
\includegraphics[width=0.451\linewidth]{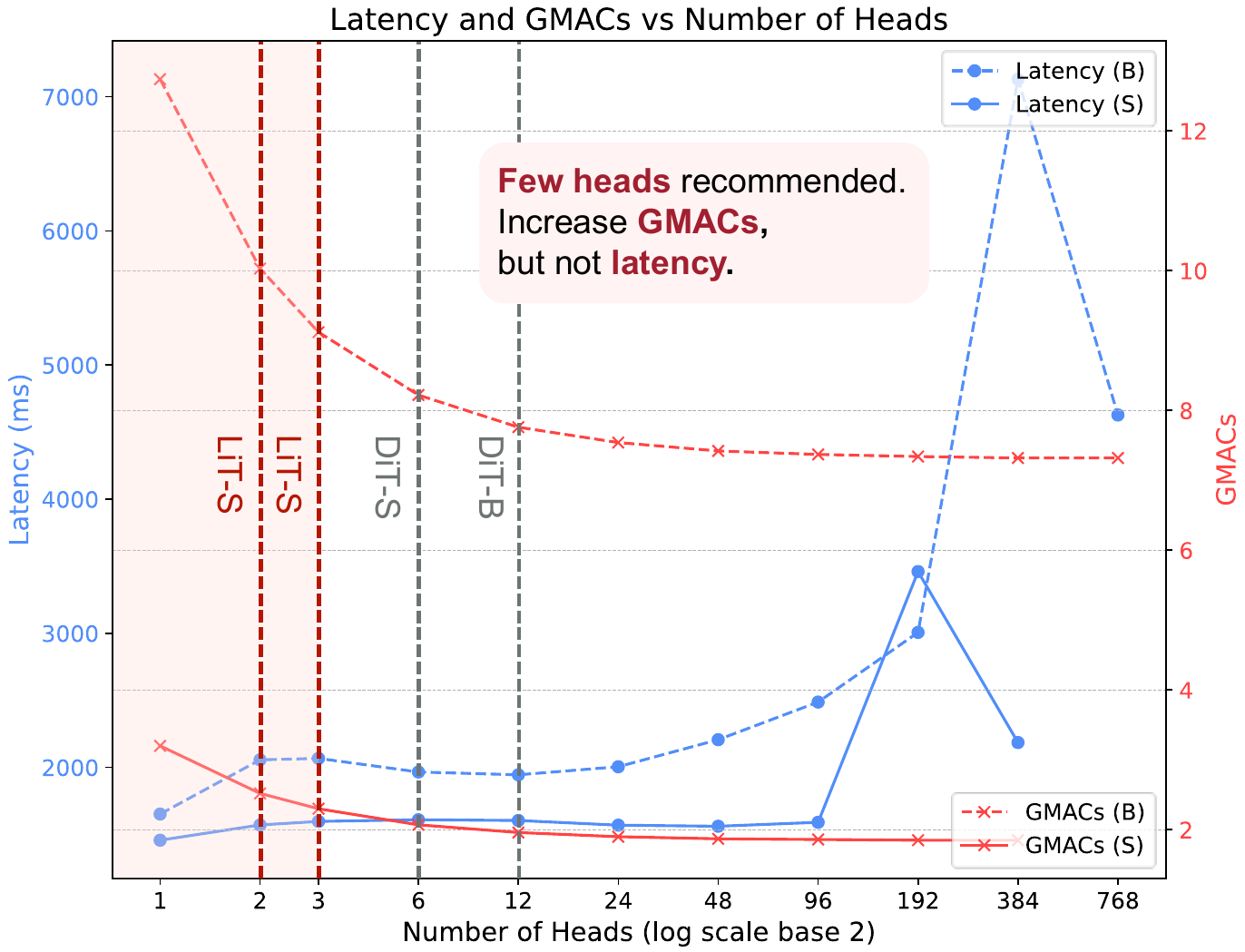} &
\includegraphics[width=0.455\linewidth]{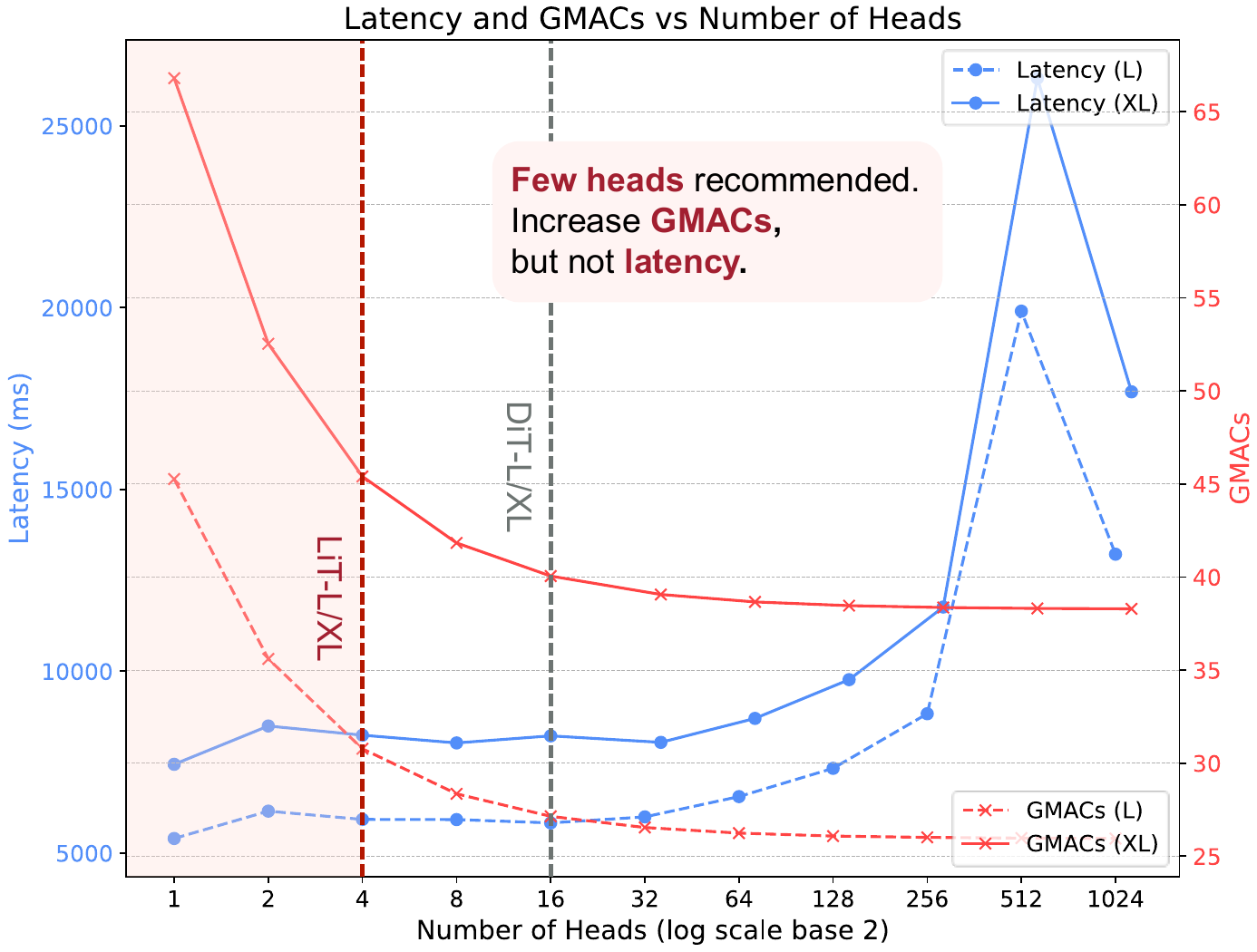} \\
(a) Small/Base & ~~~ (b) Large/XLarge  \\
\end{tabular}
\vspace{-0.2em}
\caption{\textbf{Free lunch in linear attention.} {\color{latency}{Latency}} and theoretical {\color{gmacs}{GMACs}} for linear attention with different number of heads. We test the latency to generate 256px resolution images using one NVIDIA A100 GPU with a batch size of 8. Results of (a) S-2/B-2 and (b) L-2/XL-2 models were averaged over 30 times. As head number decreases, GMACs consistently increase, but no ascending trend in latency has been observed. Based on the observation, we use few heads (\eg, 2) to get the free lunch in linear attention. 
}
\label{fig:free_lunch}
\vspace{-0.4em}
\end{figure}

\paragraph{Guideline 2: Using few heads in the linear attention increases computation but not latency.} In practice, DiT-S/B/L/XL are designed to have 6/12/16/16 heads, respectively. Denoting the number of heads, sequence length, and hidden dimension as $h, N, D$, GMACs of multi-head self-attention and linear attention are $\mathcal{C}_{\mathrm{MHSA}}$ and $\mathcal{C}_{\mathrm{MHLA}}$, respectively, as follows:
\begin{equation}
\small
\begin{aligned}
\mathcal{C}_{\mathrm{MHSA}} &=\underbrace{4ND^2}_{\rm Proj}\ +\ \underbrace{2N^2D}_{\rm Self\ Attention}\ ,\\
\mathcal{C}_{\mathrm{MHLA}} &=\underbrace{4ND^2}_{\rm Proj}\ +\ \underbrace{ND+3ND^2/h}_{\rm Linear\ Attention}\ +\  \underbrace{k^2ND}_{\rm DWC},
\end{aligned}
\label{equ:method_1}
\end{equation}
Theoretically, $\mathcal{C}_{\mathrm{MHSA}}$ is independent of the number of heads, but in linear attention, $\mathcal{C}_{\mathrm{MHLA}}$ has a negative correlation with it.
% Accordingly, in visual recognition, hydra attention~\cite{bolya2022hydra} uses as \textit{many} heads as possible in the linear attention to achieve computationally efficient. 
Intuitively, using \textit{many} heads can loose the computational stress. 
But on the contrary, we prefer to use \textit{few} heads, based on the identified \textit{free lunch} effect of linear attention. 
We illustrate the effect in Fig.~\ref{fig:free_lunch}, where we visualize the {\color{latency}{latency}} and theoretical {\color{gmacs}{GMACs}} of linear attention in DiT-S/B/L/XL models with different head numbers. We use one NVIDIA A100 GPU to generate 256$\times$256 resolution images with a batch size of 8, averaging the results over 30 experiments. 
The results show that \textit{decreasing the number of heads leads to a steady increase in theoretical {\color{gmacs}{GMACs}} but not the practical {\color{latency}{latency}}, which even shows a decrease.}
Accordingly, we summarize it as the free lunch effect of linear attention. 
We also observe this phenomenon on NVIDIA V100 GPU, as reported in Fig.~3 in Appendix Sec.~C.
As such, we argue that setting a small number of heads enables a high theoretical compute budget, which, according to scaling law~\cite{liang2024scaling, kaplan2020scaling, hoffmann2022training}, allows the model to reach a higher upper bound in generation performance.
The argument can be empirically demonstrated in Tab.~\ref{table-4-2}. 
For different model scales, equipping linear attention with a few heads (\eg, 2, 3, 4) outperforms the default settings in DiT. Instead, using an excessive number of heads (\eg, 96 for S/2 or 192 for B/2) seriously hinders the generation quality.

Besides, we also evaluate the necessity of using many heads in linear attention from a \textit{representation similarity} lens. Specifically, we visualize the average cosine similarity among the attention maps of different heads in LiT-S/2 (with 6 heads). As depicted in Fig.~4 in Appendix Sec.~D, for different denoising steps or layers, the attention across different heads is quite homogeneous, with the average cosine similarity reaching up to above 0.5. The observation suggests that a small number of heads dominate the major information in the linear attention, which positively supports the empirical generative results in Tab.~\ref{table-4-2}.

% 在实践中，DiT中的self-attention一般设计为多个头。比如，DiT-S/B/L/XL分别设置了6/12/16/16。定义heads数量，序列长度，hidden dimension分别为h,N,D,self-attention和linear attention的计算复杂度可以表示为：
% 理论上，S 与heads的数量无关。但是这种情况并不适用于linear attention。
% Bolya etal 提出hydra trick，即在分类模型的 linear attention 中使用尽可能多的heads来追求计算高效。
% 与这种做法相反，我们倾向于使用更少的heads，基于所观察的线性注意力的 ''免费的午餐"。
% 我们将线性注意力免费的午餐可以总结为：随着head数量的减少，理论GMACs不断上升，但是实际GPU上的延时并不增加 (甚至出现明显下降)。这一观察可以通过图2illustraed，我们可视化了DiT-S/B/L/XL模型使用不同heads时线性注意力的时延和理论GMACs。我们使用1 NVIDIA A100 GPU，生成 256×256 分辨率图片with batch size为8，并将结果取30次实验的均值。
% 基于这一观察，通过设置很少的heads数量，我们就可以获得更高的理论compute budget。Accordingly,缩放定律的理论基础保证了the generation performances。
% 我们进一步通过表2的结果实证性地证明了这一点。对于不同规模的模型，few heads (比如2，3)都获得最佳的性能。使用过量heads (比如对S/2使用96或对B/2使用192) 反而会使得模型无法有效生成。
% 之前的工作讨论了对用于视觉识别任务的Transformer中的self-attention使用单个heads。我们第一个在用于生成式扩散模型的linear attention研究这个问题，并清晰地对比了主流硬件设备的时延，计算量和生成性能。
% 我们最终将P中所有linear attention的 head数设置为2，因其对更大的模型友好，并在后续实验中维持这个选择。
% 除了上述免费的午餐视角，我们也从表征相似度的角度研究多头的线性注意力的必要性。图3可视化了PixArt (6 heads) 不同heads的注意力矩阵之间的平均余弦相似度。结果显示对不同的denoising steps或者层 index，不同heads的注意力很接近 (余弦相似度在0.5以上)。这些结果表明很少的heads可以表达注意力矩阵的主要信息，与表2的实证性生成结果结论相符。

\begin{figure}[t]
	\centering
	\includegraphics[width=0.9\linewidth]{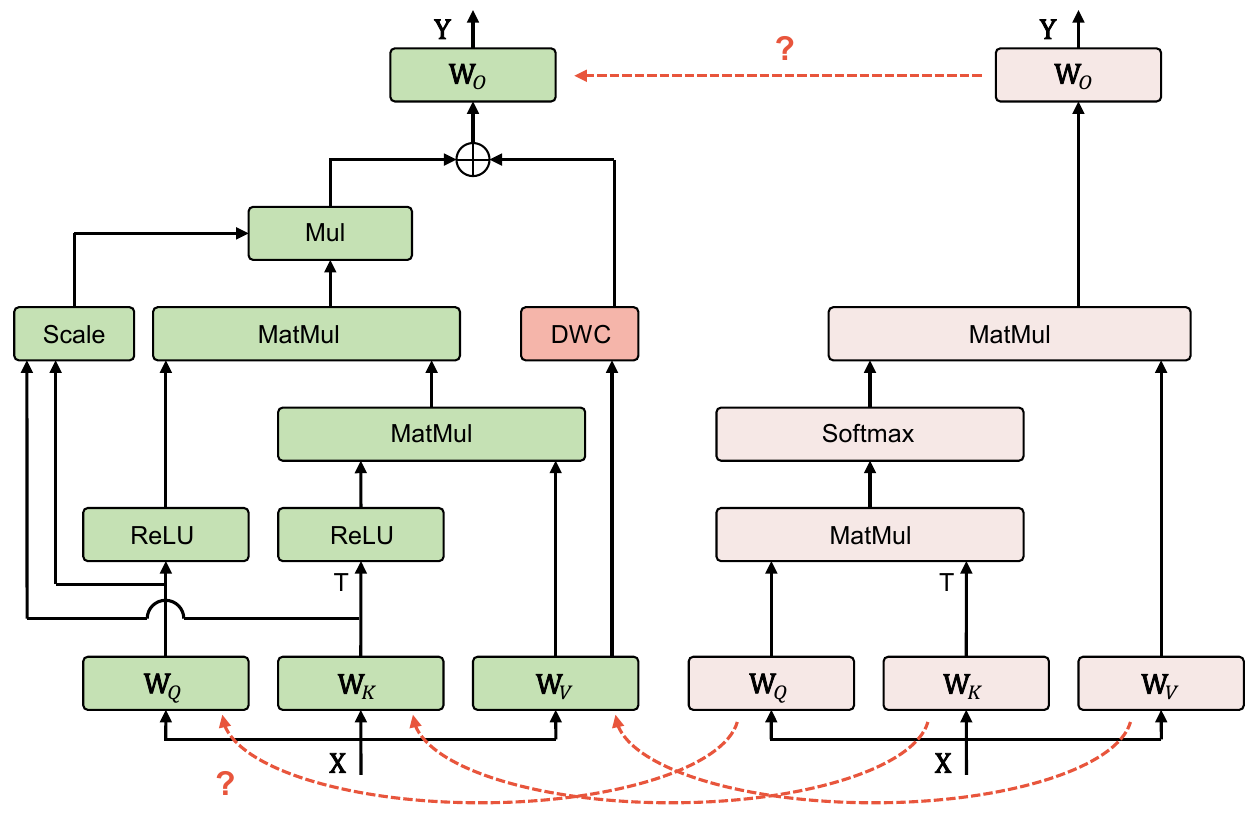}
    \vspace{-0.2em}
	\caption{\textbf{Should linear attention weights inherit from full attention?} We explore whether this should be done in Sec.~\ref{sec:method.3}.}
	\label{fig:init_g4}
    \vspace{-0.2em}
\end{figure}

\noindent \textbf{Remark 2.} Previous works~\cite{michel2019sixteen, yun2024shvit} discuss the a single head self-attention for ViTs for classification. We first study such practice of linear attention in diffusion models, providing a clear comparison of latency on mainstream GPU, GMACs, and generation performance. 
% For different model capacities (S/B/L/XL), we set the number of heads for linear attention in LiT to 2/3/4/4 for subsequent class-conditional ImageNet experiments. 

\subsection{Inheriting Weights from DiT}
\label{sec:method.3}

LiT can be converted from a pre-trained DiT, whose weights contain rich knowledge of noise prediction, enabling a cost-effective transfer to LiT. 
Thus, we load the pre-trained DiT weights into LiT, except for the linear attention. 
DiT can be pretrained to varying convergence levels. Given the architectural differences between DiT and LiT, whether converged DiT weights serve as optimal initialization for LiT remains an open question. To this end, we conduct ablation experiments to explore \textit{how the degree of training on inherited weights affects the model's generation performance}.
% Loaded weights includes FFN, adaptive layer norm (adaLN), positional encoding, and conditional embedding. 

\vspace{-0.1in}

% PixArt baseline除了线性注意力机制之外的其他部分与DiT一致，因此这些部分的参数可以从预训练的其中继承。这些部分的权重包含了与噪声预测相关的丰富的知识，被期望迁移到PixArt中。
% 预训练的DiT与PixArt的架构不完全一致。因此，对于关键的模块 (比如FFN) ，在DiT上充分训练的权重可能未必适合PixArt。因此，我们进行消融实验探索继承的权重的训练程度对于模型生成性能的影响。

\paragraph{Guideline 3: Linear diffusion Transformer should be initialized from a converged DiT.} We first pre-train DiT-S/2~\cite{peebles2023scalable} for five types of iterations—200K, 300K, 400K, 600K, and 800K—and in separate experiments, load these pre-trained weights into LiT-S/2, keeping only the linear attention parameters randomly initialized.
The initialized LiT-S/2 is then trained for 400K iterations on class-conditional ImageNet, with results shown in Tab.~\ref{table-4-3}. 
Some interesting findings are identified: (1) DiT’s pre-trained weights, even trained only 200K steps, play a significant role, improving the FID from 63.24 to 57.84. (2) Using exponential moving average (EMA)~\cite{polyak1992acceleration} of pre-trained weights has minimal impact. (3) Longer pre-training of DiT makes them better suited as initialization for LiT, even though the architectures are not fully aligned.

\noindent \textbf{Remark 3.} We suppose a possible explanation for this phenomenon is the functional decoupling of different modules in DiTs, as explored in large language models~\cite{yao2024knowledge, wang2024wise}. Basically, although DiT and LiT have different architectures, their shared components (\eg, FFN and adaLN) act quite similarly. As a result, knowledge in these components can be transferred, and extensive pre-training may not hinder the optimization of the non-transfered parts. 

\vspace{-0.1in}

\begin{table}
    \begin{center}
        \small
        \begin{tabular}{lccccc}
            \hline
             Iter. & Teacher & $\lambda_1$ & $\lambda_2$ & FID-50K ($\downarrow$) & IS ($\uparrow$) \\
            \hline

            800K & DiT-S/2 & 0.1 & 0.0 & 55.11 & 26.28 \\

            800K & DiT-XL/2 & 0.0 & 0.0 & \underline{53.83} & \underline{27.16} \\
            
            800K & DiT-XL/2 & 0.1 & 0.0 & 53.05 & 27.43 \\

            800K & DiT-XL/2 & 0.05 & 0.0 & 53.41 & 27.26 \\
            
            \gr
            
            800K & DiT-XL/2 & 0.5 & 0.0 & 51.13 & 28.89 \\

            \hline
            
            800K & DiT-XL/2 & 0.1 & 0.05 & 52.76 & 27.70 \\

            800K & DiT-XL/2 & 0.0 & 0.05 & 53.49 & 27.26 \\

            800K & DiT-XL/2 & 0.05 & 0.05 & 53.14 & 27.46 \\

            \gr
            
            800K & DiT-XL/2 & 0.5 & 0.05 & \textbf{50.79} & \textbf{29.17} \\

            \hline
        \end{tabular}
    \end{center}
    \vspace{-0.2in}
    \caption{\textbf{Knowledge distillation results on ImageNet 256$\times$256}. Baseline linear DiT-S/2 (using 2 heads), initialized using pretrained DiT-S/2, are trained for 400K steps. \underline{w/o distillation} .}
    \label{table-4-4}
    \vspace{-0.2in}
\end{table}

\paragraph{Guideline 4: Projection matrices of query, key, value, and output in linear attention should be initialized randomly.} Linear attention shares the query, key, value, and output projection matrices with full attention. Despite differing computation paradigms, these weights can be directly transferred from DiT to LiT without modification. 
However, as shown in Fig.~\ref{fig:init_g4}, it remains an open question whether this can accelerate convergence. We use a DiT-S/2 pre-trained for 600K iterations for ablation. 
We explore five different loading strategies, \ie, 
(1) $\mathbf{Q}$, $\mathbf{K}$, and $\mathbf{V}$ projection matrices; (2) $\mathbf{K}$, and $\mathbf{V}$ projection matrices; (3) $\mathbf{V}$ projection matrix; (4) $\mathbf{Q}$ projection matrix; and (5) the output projection matrix. 
Evaluation results are presented in Tab.~\ref{table-4-3}. 
Compared to the baseline without self-attention weight loading, none of the explored strategies improve generative performance. This can be attributed to differences in computational paradigms: linear attention directly computes the product of key and value matrices, unlike self-attention. As a result, self-attention's key and value weights provide limited benefit to linear attention. As shown in Tab.~\ref{table-4-3}, inheriting these weights leads to greater performance degradation, with FID scores of 55.29 and 55.07.

\noindent \textbf{Remark 4.} Inheriting weights from a LiT-like pre-trained model is a favorable approach. We recommend loading all pre-trained parameters except for the linear attention, as it is easy to implement and well-suited for diffusion models based on the Transformer macro architecture. With substantial GPU resources, extensive pre-training can further improve linear LiT's convergence and optimization, resulting in an efficient but strong model.

\begin{table}[t]
    \begin{center}
    \begin{small}
    \scalebox{0.823}{
    \begin{tabular}{lccccc}
    \toprule
    \multicolumn{6}{l}{\bf{Class-Conditional ImageNet} 256$\times$256} \\
    \toprule
    Model &   Training Steps    & FID$\downarrow$   & IS$\uparrow$   & Prec.$\uparrow$ & Rec.$\uparrow$ \\
    \toprule
    
    \dit{} DiT-S/2~\cite{peebles2023scalable} & 400K   & 68.40 & - & - & - \\
    \dit{} DiT-B/2      & 400K        & 43.47 & - & - & - \\
    \dit{} DiT-L/2      & 400K        & 23.33 & - & - & - \\
    \dit{} DiT-XL/2     & 400K         & 19.47 & - & - & - \\

    \toprule
    
    \mamba{} DiG-S/2~\cite{zhu2024dig} & 400K   & 62.06 & 22.81 & 0.39 & 0.56 \\
    \mamba{} DiG-B/2~\cite{zhu2024dig} & 400K   & 39.50 & 37.21 & 0.51 & 0.63 \\
    \mamba{} DiG-L/2~\cite{zhu2024dig} & 400K   & 22.90 & 59.87 & 0.60 & 0.64 \\
    \mamba{} DiG-XL/2~\cite{zhu2024dig} & 400K   & 18.53 & 68.53 & 0.63 & 0.64 \\
    % \dit{} DiT-XL/2     & 7M         & 9.62 & 121.50 & 0.67 & 0.67 \\
    
    \arrayrulecolor{gray}\cmidrule(lr){1-6}

    {\color{orange}$\bigstar$} \textbf{LiT-S/2} &  100K  & 60.91 & 23.51 & 0.399 & 0.583 \\
    {\color{orange}$\bigstar$} \textbf{LiT-B/2} &  100K  & 38.39 & 38.59 & 0.516 & 0.627 \\
    {\color{orange}$\bigstar$} \textbf{LiT-L/2} &  100K  & 17.80 & 76.26 & 0.639 & 0.636 \\
    {\color{orange}$\bigstar$} \textbf{LiT-XL/2} &  100K  & 12.90 & 95.80 & 0.657 & 0.653 \\
    {\color{orange}$\bigstar$} \textbf{LiT-S/2} &  400K  & 50.79 & 29.17 & 0.443 & 0.618 \\
    {\color{orange}$\bigstar$} \textbf{LiT-B/2} &  400K  & 29.55 & 50.57 & 0.557 & 0.645 \\
    {\color{orange}$\bigstar$} \textbf{LiT-L/2} &  400K  & 13.73 & 90.53 & 0.656 & 0.650 \\
    {\color{orange}$\bigstar$} \textbf{LiT-XL/2} &  400K  & 10.67 & 110.95 & 0.668 & 0.667 \\
    
    \bottomrule
    \end{tabular}
    } %
    \end{small}
    \end{center}
    \vspace{-0.2in}
    \caption{\textbf{Comparison with DiT for different model variants on class-conditional image generation.} LiT trained for only 100K steps, outperforms DiT with 400K training steps.}
    \vspace{-0.2in}
    \label{table-5-2}
\end{table}

\subsection{Distilling Both Predicted Noise and Variance}
\label{sec:method.4}

\paragraph{Guideline 5: Hybrid distillation is neccessary for student linear diffusion Transformer. We distill not only the predicted noise but also variances of the reverse diffusion process, but in a moderate way.} Knowledge distillation~\cite{hinton2015distilling} commonly employs a teacher network to help the training of a lightweight student network, which has been validated in \textit{ConvNets}~\cite{gou2021knowledge} and \textit{ViTs}~\cite{wang2023riformer, touvron2021training, zhang2022minivit}. 
For diffusion models, distillation typically focuses on reducing the sampling steps of the target model~\cite{yin2024one, yin2024improved, salimans2022progressive}. In contrast, we focus on how a heavy teacher model can aid the learning of an efficient student from an architecture perspective, while maintaining the sampling steps~\cite{liu2024linfusion}. 

A pre-trained DiT-S/2/XL/2 serves as the teacher network to distill the student LiT-S/2. We use a DiT-S/2 pre-trained for 800K steps for weight initialization of LiT, following \textbf{Guideline 3\&4}.
Denote $\mathbf{x}_t\in\mathbb{R}^{B\times C\times H\times W}$ as the noised input latent in timestep $t$, where $H$, $W$, $C$ are height, width and the number of channels of the input latent. 
$\theta^{(\mathcal{T})}, \theta^{(\mathcal{S})}$ denote network parameters of the teacher diffusion Transformer $\mathcal{T}$ and the student LiT network $\mathcal{S}$, respectively. As denoising diffusion models, the output of DiT and LiT include both noise predictions $\boldsymbol{\epsilon}_{\theta}(\mathbf{x}_t, t)$ and variances of the reverse diffusion process $\boldsymbol{\Sigma}_{\theta}(\mathbf{x}_t, t)$. DDPM~\cite{ho2020denoising} fixes the variance and optimizes model parameters using $\mathcal{L}_{simple}$. Naturally, a straightforward idea is to align the student network’s noise prediction capabilities with those of the teacher through a mean squared error (MSE) objective: 
\begin{equation}
\small
\begin{aligned}
\mathcal{L}_{noise}=\alpha_1\Vert \boldsymbol{\epsilon}_{\theta^{(\mathcal{T})}}(\mathbf{x}_t, t) - \boldsymbol{\epsilon}_{\theta^{(\mathcal{S})}}(\mathbf{x}_t, t) \Vert_F^2,
\end{aligned}
\label{equ:method_2}
\end{equation}
where $\alpha_1=1/BCHW$ is the normalization factor. The impact of distilling the predicted noise is reported in Tab.~\ref{table-4-4}. 
It shows that using a stronger teacher model (\eg, DiT-XL/2) improves the student model's generation capbility (51.13 vs. 53.83, measured by FID). Moreover, increasing the weight of the distillation objective from 0.05 to 0.5 consistently boosts performance, suggesting that LiT can benefit from the supervision of teacher DiT.

\begin{table}[t]
    \begin{center}
    \begin{small}
    \scalebox{0.809}{
    \begin{tabular}{lcccc}
    \toprule
    \multicolumn{5}{l}{\bf{Class-Conditional ImageNet} 256$\times$256} \\
    \toprule
    Model & FID$\downarrow$   & IS$\uparrow$   & Precision$\uparrow$ & Recall$\uparrow$ \\
    \toprule
    {\color{green}$\blacksquare$} BigGAN-deep~\cite{brock2019large} & 6.95 & 171.4 & 0.87 & 0.28 \\
    {\color{green}$\blacksquare$} StyleGAN-XL~\cite{sauer2022stylegan} & 2.30 & 265.12 & 0.78 & 0.53 \\
    \arrayrulecolor{gray}\cmidrule(lr){1-5}

    \unetdot{} ADM~\cite{dhariwal2021diffusion} & 10.94 & 100.98 & 0.69 & 0.63 \\
    % \unetdot{} ADM-U & 7.49 & 127.49 & 0.72 & 0.63 \\
    \unetdot{} ADM-G & 4.59 & 186.70 & 0.82 & 0.52 \\
    \unetdot{} ADM-G, ADM-U & 3.94 & 215.84 & 0.83 & 0.53 \\
    
    % \arrayrulecolor{gray}\cmidrule(lr){1-5}
    % CDM~\cite{ho2021cascaded} & 4.88 & 158.71 & - & - \\

    % LDM-8~\cite{rombach2022high} & 15.51 & 79.03 & 0.65 & 0.63 \\
    % \unetdot{} LDM-8-G & 7.76 & 209.52 & 0.84 & 0.35 \\
    % LDM-4 & 10.56 & 103.49 & 0.71 & 0.62 \\
    \unetdot{} CDM~\cite{ho2022cascaded} & 4.88 & 158.71 & - & - \\
    \unetdot{} RIN~\cite{jabri2022scalable} & 3.42 & 182.0 & - & - \\
    \unetdot{} LDM-4-G (cfg=1.25)~\cite{rombach2022high} &  3.95 & 178.22 & 0.81 & 0.55 \\
    \unetdot{} LDM-4-G (cfg=1.50) & 3.60 & 247.67 & 0.87 & 0.48 \\
    \unetdot{} Simple Diffusion (U-Net)~\cite{hoogeboom2023simple}       & 3.76 & 171.6 & - & - \\
    
    \arrayrulecolor{gray}\cmidrule(lr){1-5}
    \dit{} Mask-GIT~\cite{chang2022maskgit} & 6.18 & 182.1 & - & - \\
    \dit{} Simple Diffusion (U-ViT, L)       & 2.77 & 211.8 & - & - \\
    % \dit{} U-ViT-L/2~\cite{bao2023all}   & 3.40 & 219.94 & 0.83 & 0.52 \\
    % \dit{} U-ViT-H/2~\cite{bao2023all}   & 2.29 & 263.88 & 0.82 & 0.57 \\
    % \dit{} U-DiT-L~\cite{tian2024u}  (cfg=1.50)   & 3.37 & 246.03 & 0.862 & 0.502 \\
    \dit{} DiT-XL/2~\cite{peebles2023scalable}       & 9.62 & 121.50 & 0.67 & 0.67 \\
    \dit{} DiT-XL/2-G (cfg=1.25) & 3.22 & 201.77 & 0.76 & 0.62 \\
    \dit{} DiT-XL/2-G (cfg=1.50) & 2.27 & 278.24 & 0.83 & 0.57 \\
    \dit{} SiT-XL~\cite{ma2024sit} (cfg=1.50)   & 2.06 & 277.50 & 0.83 & 0.59 \\

    \arrayrulecolor{gray}\cmidrule(lr){1-5}
    \mamba{} DiM-L~\cite{teng2024dim}   & 2.64 & - & - & - \\
    \mamba{} DiM-H~\cite{teng2024dim}   & 2.40 & - & - & - \\
    \mamba{} DiffuSSM-XL-G~\cite{yan2024diffusion}   & 2.28 & 259.13 & 0.86 & 0.56 \\
    
    \arrayrulecolor{gray}\cmidrule(lr){1-5}
    {\color{orange}$\bigstar$} \textbf{LiT-XL/2} & 10.24 & 114.79 & 0.666 & 0.674 \\
    {\color{orange}$\bigstar$} \textbf{LiT-XL/2-G} (cfg=1.25)  & 3.60 & 191.06 & 0.758 & 0.623 \\
    {\color{orange}$\bigstar$} \textbf{LiT-XL/2-G} (cfg=1.50)  & 2.32 & 265.20 & 0.824 & 0.574 \\

    \bottomrule
    \end{tabular}
    } %
    \end{small}
    \end{center}
    \vspace{-0.2in}
    \caption{\textbf{System-level comparison on class-conditional image generation on ImageNet 256$\times$256 benchmark.} LiT-XL/2 achieves highly competitive FID using cheap linear attention.}
    \vspace{-0.15in}
    \label{table-5-3}
\end{table}

\begin{table}[t]
    \begin{center}
    % \vspace{-0.2in}
    \begin{small}
    \scalebox{0.809}{
    \begin{tabular}{lcccc}
    \toprule
    \multicolumn{5}{l}{\bf{Class-Conditional ImageNet} 512$\times$512} \\
    \toprule
    Model & FID$\downarrow$   & IS$\uparrow$   & Precision$\uparrow$ & Recall$\uparrow$ \\
    \toprule
    {\color{green}$\blacksquare$} BigGAN-deep~\cite{brock2019large} & 8.43 & 177.90 & 0.88 & 0.29 \\
    {\color{green}$\blacksquare$} StyleGAN-XL~\cite{sauer2022stylegan} & 2.41 & 267.75 & 0.77 & 0.52 \\
    \arrayrulecolor{gray}\cmidrule(lr){1-5}

    \unetdot{} ADM~\cite{dhariwal2021diffusion} & 23.24 & 58.06 & 0.73 & 0.60 \\
    \unetdot{} ADM-U & 9.96 & 121.78 & 0.75 & 0.64 \\
    \unetdot{} ADM-G & 7.72 & 172.71 & 0.87 & 0.42 \\
    \unetdot{} ADM-G, ADM-U & 3.85 & 221.72 & 0.84 & 0.53 \\
    
    % \arrayrulecolor{gray}\cmidrule(lr){1-5}
    % CDM~\cite{ho2021cascaded} & 4.88 & 158.71 & - & - \\

    % LDM-8~\cite{rombach2022high} & 15.51 & 79.03 & 0.65 & 0.63 \\
    % \unetdot{} LDM-8-G & 7.76 & 209.52 & 0.84 & 0.35 \\
    % LDM-4 & 10.56 & 103.49 & 0.71 & 0.62 \\
    % \unetdot{} CDM~\cite{ho2022cascaded} & 4.88 & 158.71 & - & - \\
    % \unetdot{} RIN~\cite{jabri2022scalable} & 3.42 & 182.0 & - & - \\
    % \unetdot{} LDM-4-G (cfg=1.25)~\cite{rombach2022high} &  3.95 & 178.22 & 0.81 & 0.55 \\
    % \unetdot{} LDM-4-G (cfg=1.50) & 3.60 & 247.67 & 0.87 & 0.48 \\
    \unetdot{} Simple Diffusion (U-Net)~\cite{hoogeboom2023simple}   & 4.28 & 171.0 & - & - \\
    
    \arrayrulecolor{gray}\cmidrule(lr){1-5}
    \dit{} Mask-GIT~\cite{chang2022maskgit} & 7.32 & 156.0 & - & - \\
    \dit{} Simple Diffusion (U-ViT, L)       & 4.53 & 205.3 & - & - \\
    % \dit{} U-ViT-L/2~\cite{bao2023all}   & 3.40 & 219.94 & 0.83 & 0.52 \\
    % \dit{} U-ViT-H/2~\cite{bao2023all}   & 2.29 & 263.88 & 0.82 & 0.57 \\
    % \dit{} U-DiT-L~\cite{tian2024u}  (cfg=1.50)   & 3.37 & 246.03 & 0.862 & 0.502 \\
    \dit{} DiT-XL/2~\cite{peebles2023scalable}       & 12.03 & 105.25 & 0.75 & 0.64 \\
    \dit{} DiT-XL/2-G (cfg=1.25) & 4.64 & 174.77 & 0.81 & 0.57 \\
    \dit{} DiT-XL/2-G (cfg=1.50) & 3.04 & 240.82 & 0.84 & 0.54 \\
    \dit{} SiT-XL~\cite{ma2024sit} (cfg=1.50)   & 2.62 & 252.21 & 0.84 & 0.57 \\

    % \arrayrulecolor{gray}\cmidrule(lr){1-5}
    % \mamba{} DiM-L~\cite{teng2024dim}   & 2.64 & - & - & - \\
    % \mamba{} DiffuSSM-XL-G~\cite{yan2024diffusion}   & 2.28 & 259.13 & 0.86 & 0.56 \\
    
    \arrayrulecolor{gray}\cmidrule(lr){1-5}
    {\color{orange}$\bigstar$} \textbf{LiT-XL/2} & 14.00 & 92.84 & 0.76 & 0.62 \\
    {\color{orange}$\bigstar$} \textbf{LiT-XL/2-G} (cfg=1.50)  & 3.69 & 207.97 & 0.85 & 0.53 \\
    \arrayrulecolor{black}\bottomrule
    \end{tabular}
    } %
    \end{small}
    \end{center}
    \vspace{-0.2in}
    \caption{\textbf{System-level comparison on class-conditional image generation on ImageNet 512$\times$512 benchmark.} LiT-XL/2 achieves highly competitive FID using cheap linear attention.}
    \vspace{-0.25in}
    \label{table-7-5}
\end{table}

As reported in IDDPM~\cite{nichol2021improved}, a suitable variance can contribute to the initial steps of the diffusion process, thereby improving the log-likelihood. Accordingly, the authors set the variance $\boldsymbol{\Sigma}_{\theta}(\mathbf{x}_t, t)$ learnable.
Based on the finding, \textit{we propose to use the variance from the teacher diffusion model as a supervision for the student, but in a moderate way}. The MSE between the reverse process variances of teacher and student can be calculated as: 
\begin{equation}
\small
\begin{aligned}
\mathcal{L}_{var}=\alpha_2\Vert \boldsymbol{\Sigma}_{\theta^{(\mathcal{T})}}(\mathbf{x}_t, t) - \boldsymbol{\Sigma}_{\theta^{(\mathcal{S})}}(\mathbf{x}_t, t) \Vert_F^2,
\end{aligned}
\label{equ:method_3}
\end{equation}
where $\alpha_2=1/BCHW$. With Eq.~\ref{equ:method_2} and Eq.~\ref{equ:method_3}, we introduce a hybrid diffusion knowledge distillation objective, involving both the predicted noise and the variance:
\begin{equation}
\small
\begin{aligned}
\mathcal{L}=\mathcal{L}_{simple} + \lambda_1\mathcal{L}_{noise} + \lambda_2\mathcal{L}_{var},
\end{aligned}
\label{equ:method_4}
\end{equation}
where $\lambda_1$, $\lambda_2$ are hyper-parameters. We thoroughly study some representative settings of our hybrid distillation objective as presented in Tab.~\ref{table-4-4}, and identifies two key findings: (1) Compared to the baseline (distilling noise only), the proposed hybrid distillation contributes positively (50.79 vs. 51.13, measured by FID), but the variance distillation should be moderate. (\eg, $\lambda_2$ set to 0.05).
(2) Distilling only the variance is counterproductive for optimization, probably as the denoising capability is the major focus for diffusion.

% 知识蒸馏一般借助一个教师模型辅助轻量化学生模型训练，已在A和B中得到验证。对于扩散模型，一般将知识从一个采样步骤较多的教师迁移到一个较少采样步骤的学生模型。我们则从网络架构角度关注复杂的教师模型对于轻量化学生学习的帮助，保持采样步数。
% 具体而言，一个预训练的DiT-XL/2作为教师模型被用来蒸馏学生PixArt-S/2。
% 可以发现当使用一个较强的教师模型 (如DiT-XL/2) 时，学生模型的fid可以有效提升 (51.13 vs 53.83)。而且，增加知识蒸馏的权重系数可以进一步改善性能，意味着简单的PixArt学生需要一个教师模型的辅助。
% 扩散模型的前几步对 VLB 很关键，而合适的方差可以借此帮助log-likelihood。
% 表4报告了使用我们的混合蒸馏目标的消融实验结果。
% 到目前为止，我们关闭了PixArt的探索过程并得到了一套基线模型和训练策略。PixArt与DiT保持宏观设计一致，但使用了更高效的线性注意力。配合我们总结的cost-effective的训练策略，与DiT-S/2的68.40的FID相比，PixArt大幅提升至50.79 (训练相同iterations)。接下来，我们在更大的模型规模 (B/L/XL)，和更复杂的任务 (文生图) 验证PixArt的有效性。

\vspace{-0.2in}

\paragraph{Closing remarks.} So far, we have finalized the roadmap with 5 practical guidelines. We firmly answer how to convert a pre-trained DiT safely and efficiently to LiT while maintaining performance. Next, we will validate it on larger capacities (\eg, B/L/XL) and text-to-image generation.

\section{Experiments}
\label{sec:experiments}

As this work focuses on \textit{safely and efficiently converting a pretrained baseline into a linear version without performance gap}, we thus choose DiT~\cite{peebles2023scalable} (for class-conditional image generation) and PixArt-$\Sigma$~\cite{chen2024pixart3} (for text-to-image generation) as baselines. The goal for LiT is to achieve comparable performance while maintaining an efficient training.

\subsection{Class-Conditional Image Generation} 
\label{sec:experiments.1}
\paragraph{256$\times$256 ImageNet.} We evaluate LiT on ImageNet~\cite{deng2009imagenet} benchmark. LiT-S/2, B/2, L/2, XL/2 configurations are consistent with DiT, except that head numbers in linear attention are 2/3/4/4, following \textbf{Guideline 2} in Sec.~\ref{sec:method.2}. DWC kernel size is set to 5 for all model variants. Detailed configuration are provided in Sec.~A of the Appendix. We train 400K steps with a batch size of 256. For LiT-XL/2, we extend training to 1.4M steps (20$\%$ of DiT's 7M steps). We used the AdamW~\cite{loshchilov2017decoupled} optimizer with a learning rate of $1e^{-4}$ and weight decay set to 0. We use 250 sampling steps, aligning with DiT. We initialize LiT’s parameters using the pre-trained DiT. $\lambda_1$, $\lambda_2$ in hybrid knowledge distillation, are set to 0.5 and 0.05. Evaluation metrics include FID-50K~\cite{heusel2017gans}, Inception Score~\cite{salimans2016improved} and Precision/Recall~\cite{kynkaanniemi2019improved}. 

As reported in Tab.~\ref{table-5-2}, LiT trained for only 100K steps already outperforms the DiT trained for 400K steps across different evaluation metrics and model variants. With extra training up to 400K steps, the model's performance continues to improve. In Tab.~\ref{table-5-3}, we compare LiT-XL/2 trained for 1.4M steps with high-performance baselines. Despite having only $20\%$ of the training steps of DiT-XL/2, LiT still competes on par with DiT (2.32 vs. 2.27). Besides, LiT competes favorably with several U-Net~\cite{ronneberger2015u} based baselines. 
The results demonstrate that our LiT, when combined with suitable optimization strategies, can be reliably used in image generation applications. 

% \input{figtext/tab_5_1}

% \vspace{-0.2in}

\paragraph{512$\times$512 ImageNet.} 

We further validate LiT-XL/2 on the 512px setting. Using the pre-trained DiT-XL/2~\cite{peebles2023scalable} as the teacher, we initialize LiT-XL/2 with its weights. For knowledge distillation, we set $\lambda_1$ and $\lambda_2$ to 1.0 and 0.05, respectively, and train LiT-XL/2 for only 700K steps ($\sim 23\%$ of DiT's~\cite{peebles2023scalable} 3M steps).
Notably, unlike DiT, which used a batch size of 256, we adopt a smaller batch size of 128, requiring twice the training steps to complete 1 epoch (\ie, processing the entire training dataset once). Aside from this, all other hyper-parameter settings were consistent with the 256$\times$256 experiments. The results are reported in Tab.~\ref{table-7-5}.
Despite being trained for only 700K steps, which is equivalent in training epochs to 350K steps with a batch size of 256 on the same dataset,  LiT, using pure linear attention, achieves an impressive FID of 3.69, comparable to DiT trained for 3M steps, reducing the training steps by $\sim 77\%$. Additionally, LiT outperforms several strong baselines. These results demonstrate the effectiveness of our proposed 5 practical guidelines on higher-resolution setup.

\subsection{Text-to-Image Generation} 
\label{sec:experiments.2}

% \vspace{-0.1in}

\begin{table}[t]
    \begin{center}
    % \vspace{-0.2in}
    \begin{small}
    \scalebox{0.645}{
    \begin{tabular}{lcccccccc}
    \toprule
    Model &    \#Params.  &  Single & Two & Count. & Colors & Pos. & Attri. & Overall  \\
    \toprule

    \unetdot{} LDM~\cite{rombach2022high} & 1.4B & 0.92 & 0.29 & 0.23 & 0.70 & 0.02 & 0.05 & 0.37 \\
    \unetdot{} SDv1.5~\cite{rombach2022high} & 0.9B & 0.97 & 0.38 & 0.35 & 0.76 & 0.04 & 0.06 & 0.43 \\
    \unetdot{} SDv2.1~\cite{rombach2022high}  & 0.9B & 0.98 & 0.51 & 0.44 & 0.85 & 0.07 & 0.17 & 0.50 \\
    \arrayrulecolor{gray}\cmidrule(lr){1-9}
    \dit{} LlamaGen~\cite{sun2024autoregressive} &  0.8B & 0.71 & 0.34 & 0.21 & 0.58 & 0.07 & 0.04 & 0.32 \\
    \dit{} PixArt-$\alpha$~\cite{chen2024pixart1} & 0.6B & 0.98 & 0.50 & 0.44 & 0.80 & 0.08 & 0.07 & 0.48 \\  
    \dit{} PixArt-$\Sigma$~\cite{chen2024pixart3} & 0.6B & - & - & - & - & - & - & 0.52 \\ 
    \dit{} Lumina-Next~\cite{zhuo2024lumina} & 2.0B & - & - & - & - & - & - & 0.46 \\
    \arrayrulecolor{gray}\cmidrule(lr){1-9}
    \mlpmixerdot{} SEED-X~\cite{ge2024seed}  & 17B & 0.97 & 0.58 & 0.26 & 0.80 & 0.19 & 0.14 & 0.49 \\
    \mlpmixerdot{} Chameleon~\cite{team2405chameleon}  & 34B & - & - & - & - & - & - & 0.39 \\
    \mlpmixerdot{} LWM~\cite{liu2024world}  & 7B & 0.93 & 0.41 & 0.46 & 0.79 & 0.09 & 0.15 & 0.47 \\
    \arrayrulecolor{gray}\cmidrule(lr){1-9}
    {\color{orange}$\bigstar$} \textbf{LiT} (512px) & 0.6B & 0.97 & 0.43 & 0.42 & 0.79 & 0.09 & 0.12 & 0.47 \\
    {\color{orange}$\bigstar$} \textbf{LiT} (1024px) & 0.6B & 0.98 & 0.50 & 0.40 & 0.77 & 0.11 & 0.12 & 0.48 \\
    \arrayrulecolor{black}\bottomrule
    \end{tabular}
    } %
    \end{small}
    \end{center}
    \vspace{-0.2in}
    \caption{\textbf{Evaluation of text-to-image generation ability on GenEval benchmark.} LiT uses linear attention in each block.}
    % \vspace{-0.25in}
    \label{table-geneval}
\end{table}

\paragraph{Experimental setup.}   
Building up with PixArt-$\Sigma$~\cite{chen2024pixart3} as the baseline, we develop LiT-0.6B with 2 linear attention heads and a DWC kernel size of 5. We leverage the pretrained SDXL~\cite{podell2023sdxl} VAE Encoder and T5~\cite{chung2024scaling} Encoder (\ie, Flan-T5-XXL) for image and text feature extraction, respectively. 
In 512px setting, we use PixArt-$\Sigma$ as the teacher to supervise the training of LiT-0.6B, with $\lambda_1$, $\lambda_2$ set to 1.0, 0.05. As in Sec.~\ref{sec:method.3}, we inherit the weights of LiT from PixArt-$\Sigma$, except for the parameters in the whole attention module. 
Subsequently, we train LiT-0.6B on an internal dataset, with a mixture of long (sharegpt caption, $60\%$) and short (original caption, $40\%$) captions. 
% Token length is increased from $120$ to $300$.
Token length is set to $300$.
We train with a learning rate of $2e^{-5}$ and batch size of 8$\times$48 for $\sim$106K steps, which is efficient compared to the multi-stage training used for PixArt-$\alpha$~\cite{chen2024pixart1}. 
This may also be related to our proposed weight inheritance strategy. 
We use 20 sampling steps, following PixArt-$\Sigma$. 
In 1K setting, we basically follow the 512px seting for training and qualitative evaluation. We choose 512px LiT-0.6B as the starting point for training, with positional embeddings interpolation applied. We use the PixArt-$\Sigma$~\cite{chen2024pixart3} for 1K resolution as the teacher model. 
Starting from the 512px LiT-0.6B, we continue training on the internal datasets for $\sim$155K steps with a batch size of 2$\times$48.
% We train with a batch size of 48$\times$2 for $\sim$155K steps on the internal datasets. 
We select suitable LiT-0.6B variants for sampling qualitative results and laptop deployment.

\paragraph{Results.} 
Tab.~\ref{table-geneval} compare the GenEval~\cite{ghosh2023geneval} results of LiT with those of other powerful methods.
Despite relying solely on linear attention, LiT competes on par with its baseline PixArt-$\Sigma$, and surpasses several strong text-to-image generation baselines, such as Lumina-Next~\cite{zhuo2024lumina}, LlamaGen~\cite{sun2024autoregressive}, and SDv1.5~\cite{rombach2022high}.
Fig.~5,~6, and~7 (Sec.~F of the Appendix) depict the 512px images generated by LiT.  
LiT-0.6B can produce exceptional, photorealistic images.
Fig.~\ref{fig:t2i-outputs-1k} shows the sampled 1K resolution images generated by LiT. Our LiT-0.6B is able to generate images at a higher resolution with fine details and colorful contents, with pure attention in each block. 
These results show the generality of our proposed practical guidelines in text-to-image frameworks.

\begin{table}[t]
    \begin{center}
    \begin{small}
    \scalebox{0.72}{
    \begin{tabular}{lccccc}
    \toprule
    Model &     \#Params.  & Attention  &  Laptop   & Latency (1K) & Latency (2K)  \\
    \toprule  
    PixArt-$\Sigma$~\cite{chen2024pixart3} & 0.6B & KV Compress & \ding{55} & 4.38s & 32.16s \\ 
    LiT & 0.6B & Linear & \ding{51} & 3.93s & 14.59s \\
    \arrayrulecolor{black}\bottomrule
    \end{tabular}
    } %
    \end{small}
    \end{center}
    \vspace{-0.2in}
    \caption{\textbf{Latency and Efficient Attention Comparison.} Latency is measured with batch size 1 and 20 sampling steps.}
    \vspace{-0.28in}
    \label{table-system-latency}
\end{table}

\vspace{-0.17in}

\paragraph{Efficiency Analysis} We compare the latency of LiT and our baseline PixArt-$\Sigma$~\cite{chen2024pixart3} for generating 1K and 2K resolution images, as shown in Tab.~\ref{table-system-latency}. The results are averaged over 10 runs on a 40GB NVIDIA A100 GPU. While PixArt-$\Sigma$ employs key-value compression~\cite{wang2022pvt} for attention acceleration, LiT's linear attention proves more efficient, reducing latency by \textbf{$\sim10\%$} at 1K and \textbf{$\sim55\%$} at 2K resolution. 

\vspace{-0.18in}

\paragraph{On-Device Offline Deployment} 

We deploy our LiT-0.6B for 1K resolution to a laptop powered by the Windows 11 operating system to validate its on-device capability. 
Specifically, we quantize the text encoder to 8 bits while using fp16 precision for other weights. 
The results (two photos) are provided in Sec.~F of the Appendix. 
LiT-0.6B generates photorealistic 1K-resolution images offline, showcasing its efficiency for edge computing.

\section{Conclusion and Limitation}
\label{sec:conclusion}

This paper answers the question of how to safely and efficiently convert a pre-trained DiT for (text-to-)image generation to a linear DiT without compromising model capacity. We reexamine attention head number, weight inheritance, and knowledge distillation, and propose 5 useful guidelines as the response. In future work, we will focus on scalability and high-resolution (\eg, 4K) LiT generation.

\section*{Acknowledgement}
This paper is partially supported by the National Key R\&D Program of China No.2022ZD0161000. 

{
    \small
    \bibliographystyle{ieeenat_fullname}
    \bibliography{main}
}

\clearpage

\appendix

\clearpage
\setcounter{page}{1}
\maketitlesupplementary

\section{Model Configuration}
\label{sec:appendix.1}

% 我们汇报了
We report the configurations of LiT variants in Tab.~\ref{table-5-1}, which basically follow the hyper-parameters of DiT~\cite{peebles2023scalable}, except for using few heads. For class-conditional image generation, we set 2/3/4/4 heads for linear attention used in LiT.
For LiT-XL/2, used for text-to-image tasks, we use 2 heads for linear attention.

\begin{table}[h]
\begin{center}
\small
% \vspace{0.02in}
\scalebox{0.97}{
\begin{tabular}{lccccc}
\toprule
Model            & Layers & Hidden size &  Heads & Patch size \\
\toprule
LiT-S   &   12   &     384   &   2    &   2  \\
LiT-B   &   12   &      768    &   3   &   2  \\
LiT-L  &    24   &      1024    &   4   &  2  \\
LiT-XL &    28  &       1152     &   4  &  2  \\
% \hline
LiT-XL$^\diamond$ &    28  &       1152     &   2  &  2  \\
\bottomrule
\end{tabular}}
\end{center}
\vspace{-0.12in}
\caption{\textbf{Configuraions of LiT} for class-conditional image generation and text-to-image generation (denoted by $^\diamond$). Apart from using few heads, we generally follow the DiT~\cite{peebles2023scalable} setting.}
\vspace{-0.12in}
\label{table-5-1}
\end{table}

\section{Latency Analysis in Diffusion Transformer}
\label{sec:appendix.2}

\begin{figure}[t]
	\centering
	\includegraphics[width=0.94\linewidth]{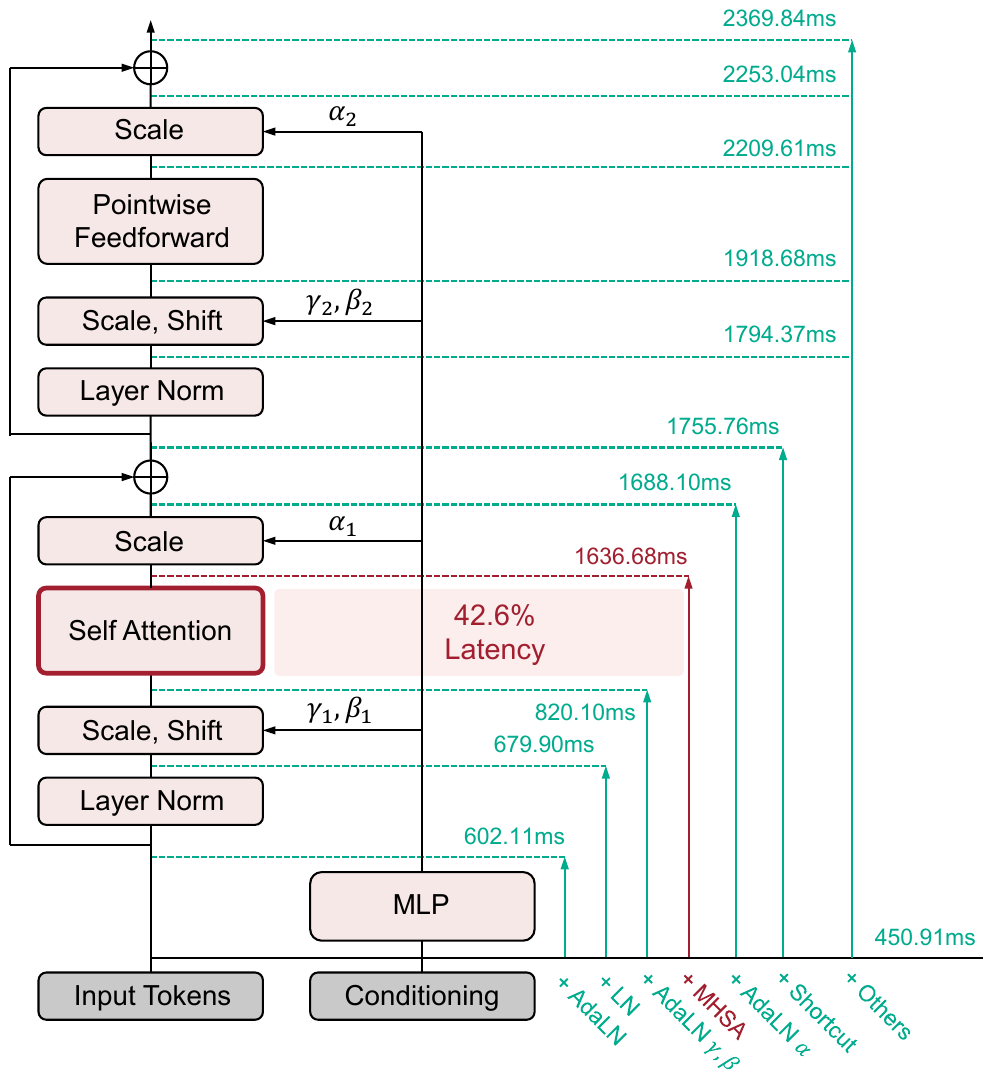}
    \vspace{-0.05in}
	\caption{\textbf{Latency analysis of different components} in DiT-B/4~\cite{peebles2023scalable} with a batch size of 8 using NVIDIA A100 GPU. Latency of the vanilla self-attention occupies about \textbf{42.6\%} of the backbone. Our LiT successfully replaces the heavy attention module with simple linear attention, by using the proposed architectural design and training guidelines.}
	\label{fig:latency}
    % \vspace{-0.2in}
\end{figure}

We conduct a component-wise latency analysis of the Diffusion Transformer, with results shown in Fig.~\ref{fig:latency}. The analysis was performed using the DiT-B/4~\cite{peebles2023scalable} model on an NVIDIA A100 GPU. The results indicate that the self-attention module accounts for 42.6\% of the total latency of a DiT block. We attribute the observed considerable latency proportion to the quadratic computational complexity of the self-attention. 

\section{Detailed Latency and Theoretical GMACs}
\label{sec:appendix.3}
We use one NVIDIA V100 GPU to evaluate the {\color{latency}{latency}} and theoretical {\color{gmacs}{GMACs}} of the DiT-S/B models with different numbers of heads. The task was to generate 256$\times$256 resolution images with a batch size of 8, following the experimental setup of Fig.~5 in the main paper (except for the GPU type), 
and the results are shown in Fig.~\ref{fig:free_lunch_v100}. 

We observe that both the small and base models on the V100 GPU exhibit a phenomenon similar to that on the A100 GPU: as theoretical {\color{gmacs}{GMACs}} increased, practical {\color{latency}{latency}} does not follow the same trend and even descend (in the B/2 model). This finding supports the generalizability of the free lunch
effect of linear attention.

\begin{figure}[t]
	\centering
	\includegraphics[width=0.9\linewidth]{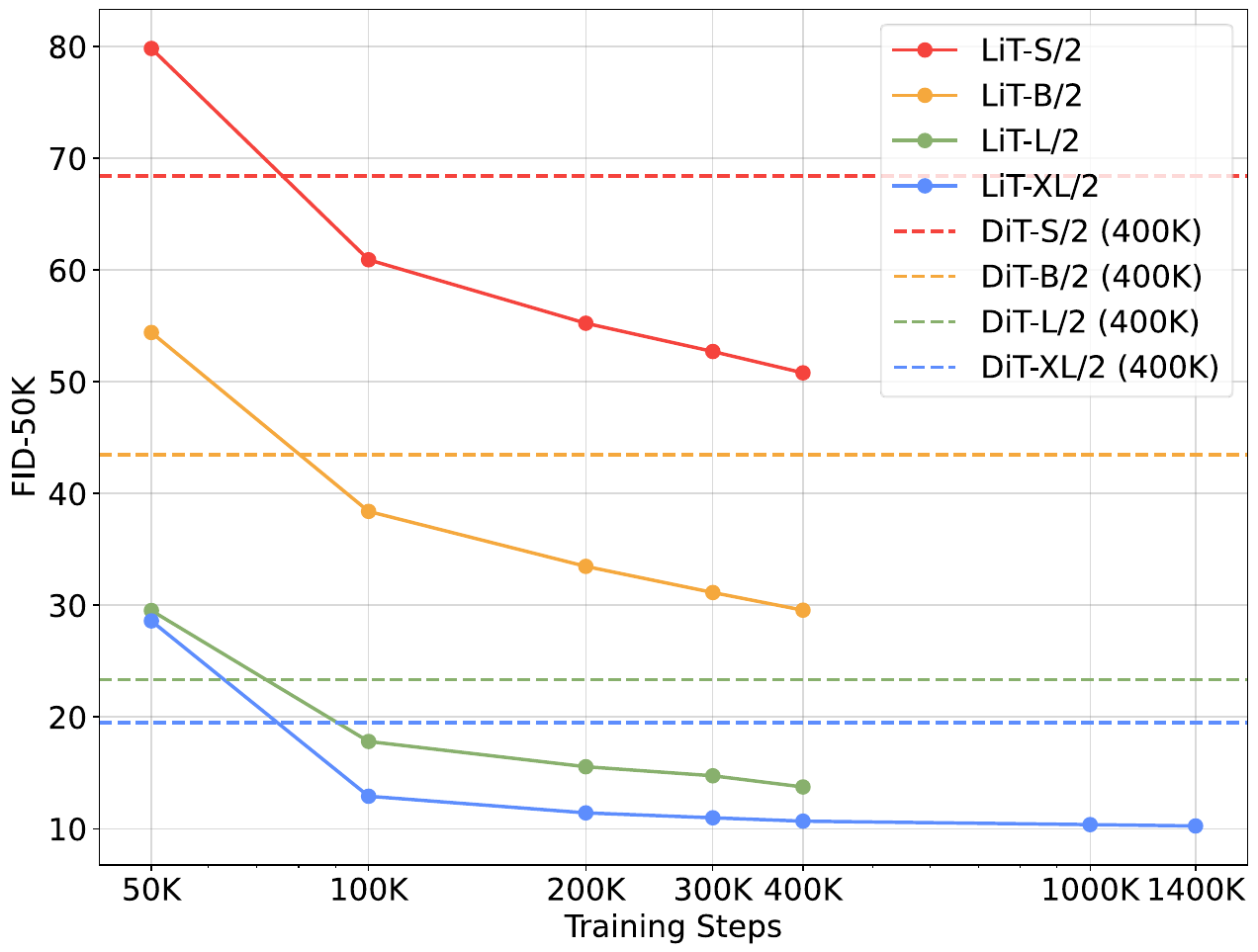}
    \vspace{-0.05in}
	\caption{\textbf{Comparing training efficiency between our LiT and DiT}. LiT outperforms DiT (400K training steps) with only 100K training steps for different model sizes.}
	\label{fig:results}
    % \vspace{-0.05in}
\end{figure}

\begin{figure}[t]
	\centering
	\includegraphics[width=1.0\linewidth]{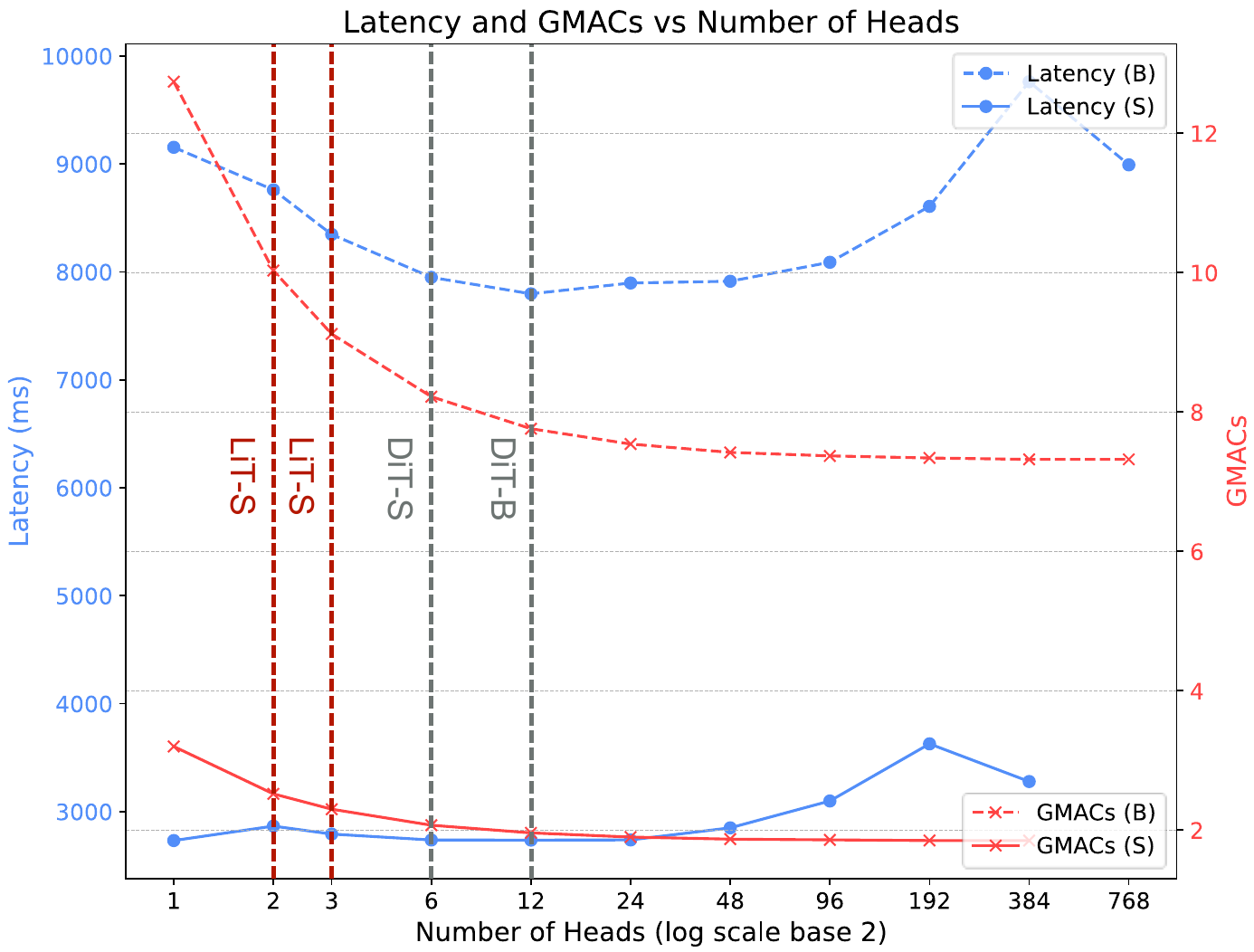}
    % \vspace{-0.2in}
	\caption{\textbf{Free lunch in linear attention.} Comparison of {\color{latency}{latency}} and theoretical {\color{gmacs}{GMACs}} for linear attention with different number of heads. We test the latency to generate $256\times256$ resolution images using one NVIDIA V100 GPU with a batch size of 8. Results of S/2 and B/2 model were averaged over 30 times. Results for the case of the V100 GPU demonstrate a similar phenomenon to the A100 GPU.}
	\label{fig:free_lunch_v100}
    % \vspace{-0.2in}
\end{figure}

\section{Detailed Results on Attention Head Similarities}
\label{sec:appendix.4}
\begin{figure}[t]
	\centering
	\includegraphics[width=0.9\linewidth]{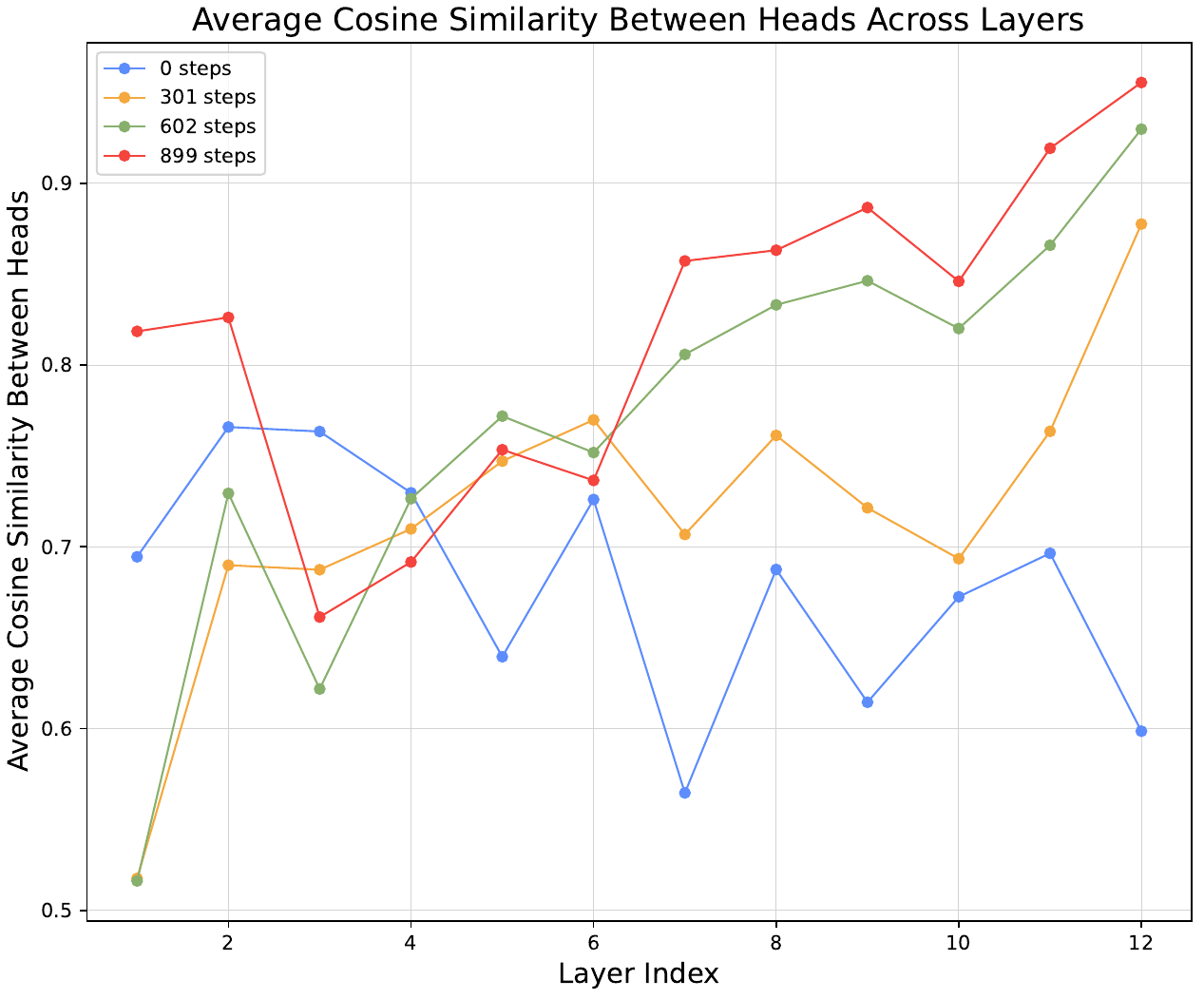}
    \vspace{-0.1in}
	\caption{\textbf{Redundancy in linear attention heads.} Attention maps of different heads of LiT-S/2 (6 heads) show high average cosine similarity.}
	\label{fig:cos_sim_head}
    \vspace{-0.0in}
\end{figure}

We take LiT-S/2 with 6 heads for visualization. Average cosine similarity among the attention maps of different heads is illustrated in Fig.~\ref{fig:cos_sim_head}.

\section{Detailed Results on Class-Conditional Image Generation}
\label{sec:appendix.5}

\paragraph{Detailed results on Sec. 4.}
We provide detailed results for Tab. 1, Tab. 2, Tab. 3 and Tab. 4 in the main text in Tab.~\ref{table-7-1}, Tab.~\ref{table-7-2}, Tab.~\ref{table-7-3}, and Tab.~\ref{table-7-4}, respectively. In each table, we report results involving FID-50K~\cite{heusel2017gans} (\textit{without} classifier-free guidance), Inception Score (IS)~\cite{salimans2016improved} and Precision/Recall~\cite{kynkaanniemi2019improved}.
IS, and Precision/Recall show results similar to FID-50K. As a result, the conclusions drawn in Sec. 4 of the main paper apply not only to metrics evaluating the distance between generated images and real images (\eg, FID-50K) but also to metrics reflecting the quality of the generated images themselves (\eg, IS).

\section{More Results on Text-to-Image Generation}
\label{sec:appendix.6}

\begin{figure}[t]
	\centering
	\includegraphics[width=0.9\linewidth]{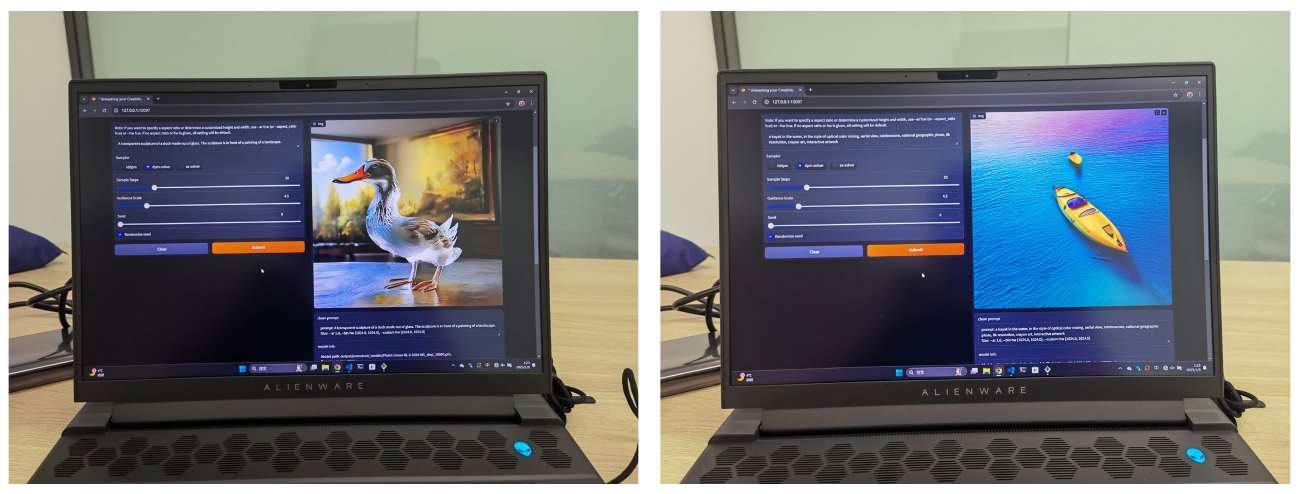}
    \vspace{-0.1in}
	\caption{\textbf{Offline deployment of on a Windows 11 laptop.} LiT runs swiftly on the edge, generating 1K resolution images.}
	\label{fig:laptop_dep}
    \vspace{-0.14in}
\end{figure}

We provide more text-to-image results in Fig.~\ref{fig:t2i-outputs2} and Fig.~\ref{fig:t2i-outputs3}. As shown, LiT can accurately generate 512px photo-realistic images in various styles, themes, and content, whether the human instructions are simple or complicated. 
These results demonstrate that LiT effectively learns useful knowledge from the teacher model while maintaining exceptional computational efficiency, validating the effectiveness of our proposed cost-effective training strategy. Results of the offline laptop development are shown in Fig.~\ref{fig:laptop_dep}.

\begin{algorithm*}[t]
\caption{Linear Attention in LiT, Pseudo-code}
\label{alg:linear}
\definecolor{codeblue}{rgb}{0.25,0.5,0.5}
\definecolor{codekw}{rgb}{0.85, 0.18, 0.50}
\lstset{
  backgroundcolor=\color{white},
  basicstyle=\fontsize{7.5pt}{7.5pt}\ttfamily\selectfont,
  columns=fullflexible,
  breaklines=true,
  captionpos=b,
  commentstyle=\fontsize{7.5pt}{7.5pt}\color{codeblue},
  keywordstyle=\fontsize{7.5pt}{7.5pt}\color{codekw},
}
\begin{lstlisting}[language=python]
import torch
import torch.nn as nn

class LinearAttention(nn.Module):
    def __init__(
        self, 
        dim, 
        num_heads=8, 
        qkv_bias=False, 
        attn_drop=0., 
        proj_drop=0., 
        kernel_function=nn.ReLU, 
        kernel_size=5,
        fp32_attention=True,
        **block_kwargs,
    ):
        super().__init__()
        assert dim % num_heads == 0, f"dim {dim} should be divisible by num_heads {num_heads}."

        self.dim = dim
        self.num_heads = num_heads
        head_dim = dim // num_heads

        self.q = nn.Linear(dim, dim, bias=qkv_bias)
        self.kv = nn.Linear(dim, dim * 2, bias=qkv_bias)
        self.attn_drop = nn.Dropout(attn_drop)
        self.proj = nn.Linear(dim, dim)
        self.proj_drop = nn.Dropout(proj_drop)

        self.dwc = nn.Conv2d(in_channels=head_dim, out_channels=head_dim, kernel_size=kernel_size,
                             groups=head_dim, padding=kernel_size // 2)
        self.kernel_function = kernel_function()
        self.fp32_attention = fp32_attention

    def forward(self, x, HW=None):
        B, N, C = x.shape
        new_N = N
        if HW is None:
            H = W = int(N ** 0.5)
        else:
            H, W = HW

        q = self.q(x)   
        dtype = q.dtype

        kv = self.kv(x).reshape(B, N, 2, C).permute(2, 0, 1, 3)   
        k, v = kv[0], kv[1]   

        q = self.kernel_function(q) + 1e-6   
        k = self.kernel_function(k) + 1e-6   

        q = q.reshape(B, N, self.num_heads, -1).permute(0, 2, 1, 3).to(dtype)   
        k = k.reshape(B, N, self.num_heads, -1).permute(0, 2, 1, 3).to(dtype)   
        v = v.reshape(B, N, self.num_heads, -1).permute(0, 2, 1, 3).to(dtype)   

        use_fp32_attention = getattr(self, 'fp32_attention', False)     
        if use_fp32_attention:
            q, k, v = q.float(), k.float(), v.float()

        with torch.cuda.amp.autocast(enabled=not use_fp32_attention):
            z = 1 / (q @ k.mean(dim=-2, keepdim=True).transpose(-2, -1) + 1e-6)   
            kv = (k.transpose(-2, -1) * (N ** -0.5)) @ (v * (N ** -0.5))    
            x = q @ kv * z   

            x = x.transpose(1, 2).reshape(B, N, C)    
            v = v.reshape(B * self.num_heads, H, W, -1).permute(0, 3, 1, 2)   
            x = x + self.dwc(v).reshape(B, C, N).permute(0, 2, 1)   

        x = x.type(torch.float16)

        x = self.proj(x)
        x = self.proj_drop(x)

        return x
        
\end{lstlisting}
\end{algorithm*}

\section{Pseudo-code}
\label{sec:appendix.7}
As presented in Alg.~\ref{alg:linear}, we provide an example of the pseudo-code for the linear attention used in our LiT model for text-to-image generation for the laptop development. 
We set the kernel size of the depthwise convolution to 5.

\section{Full Related Work}
\label{sec:appendix.8}

\paragraph{Linear attention.}
As a computationally efficient alternative to self-attention, linear attention~\cite{katharopoulos2020transformers} reduces computational complexity from quadratic to linear and has been proven effective in both visual understanding domain~\cite{cai2023efficientvit, guo2024slab, han2023flatten, han2023agent, bolya2022hydra} and language domain~\cite{qin2022cosformer, yang2023gated, qin2024hierarchically, qin2024hgrn2, chen2024dijiang}. 
EfficientViT~\cite{cai2023efficientvit} introduces a multi-scale linear attention with hardware-efficient operations to obtain a general vision backbone. Flatten Transformer~\cite{han2023flatten} introduces focused linear attention to address the deficiencies in focus ability and feature diversity of linear attention, incorporating a focused function and depthwise convolution (DWC). 
SLAB~\cite{guo2024slab} simplifies focused linear attention by retaining only the DWC component and introduces a progressive re-parameterized batch normalization to adapt offline batch normalization~\cite{ioffe2015batch} for achieving low inference latency.
These studies have been validated on visual perception tasks. Meanwhile, our work refines a linear attention module tailored for image generation tasks and identifies the free lunch of using few heads.

\paragraph{Efficient diffusion Transformer for image generation.}
Limited by the quadratic computational complexity of self-attention, recent studies focus on developing efficient Transformer-style architectures for diffusion models. 
For example, 
DiM~\cite{teng2024dim}, ZigMa~\cite{hu2024zigma}, and DiMSUM~\cite{phung2024dimsum} explore implementing Mamba-based~\cite{gu2023mamba, dao2024transformers} DiT-style~\cite{peebles2023scalable} models. 
Diffusion-RMKV~\cite{fei2024diffusion} studies RWKV-style~\cite{peng2023rwkv} models for diffusion. 
Mediator~\cite{pu2024efficient} introduces an attention mediator to obtain an efficient diffusion Transformer with linear complexity. 
DiG~\cite{zhu2024dig} replaces the self-attention in DiT with gated linear attention to speed up training. 
LinFusion~\cite{liu2024linfusion} and Sana~\cite{xie2024sana} apply linear attention to U-Net-based~\cite{ronneberger2015u} and Transformer-based~\cite{vaswani2017attention} diffusion models, respectively, and train these models from scratch to generate high-quality images based on user instructions. 
Other studies~\cite{li2024q, li2023q} explore efficient diffusion models through perspectives of low-bit quantization~\cite{han2015deep}, feature map reusing~\cite{ma2024deepcache, li2024distrifusion}, and lightweight architecture design~\cite{zhao2023mobilediffusion}. 
Differently, our work not only refines the design of linear attention but also introduces cost-effective training strategies, including weight inheritance and a novel hybrid diffusion distillation approach.

\paragraph{Advanced training method for diffusion models.}
Some studies explores improved training strategies to enhance the optimization of diffusion models or achieve more efficient models. For example, 
CAN~\cite{cai2024condition} introduces a condition-aware weight generation module to the diffusion Transformer, and demonstrate the technique can be further equipped with EffcientViT~\cite{cai2023efficientvit} to achieve both effectiveness and efficiency. 
REPA~\cite{yu2024representation} proposes aligning the intermediate features of the denoising model with those extracted by a pre-trained visual encoder during the training of diffusion models.
Some studies~\cite{salimans2022progressive, yin2024one, yin2024improved} explore distillation techniques to reduce the sampling steps of the diffusion model. Unlike the goal of reducing sampling steps, our proposed hybrid knowledge distillation focuses on an architectural perspective, exploring how a lightweight student model can learn from a computationally intensive teacher model.

\begin{table*}
    \begin{center}
        \small
        \begin{tabular}{lcccccccc}
            \hline
            DiT & Attention  & Resolution & Batch Size & Training Steps & FID-50K ($\downarrow$) & IS ($\uparrow$) & Precision ($\uparrow$) & Recall ($\uparrow$)	\\
            \hline
            S/2    & softmax & 256 & 256 & 400K & 68.40 & -  \\
            
            S/2    & ReLU linear & 256 & 256 & 400K & 88.46 & 15.11 & 0.29 & 0.45 \\

            \gr
            S/2    & Simplified linear (ReLU) & 256 & 256 & 400K & 63.66 & 22.16 & 0.38 & 0.58 \\
            
            S/2    & focused linear (ReLU) & 256 & 256 & 400K & \textbf{63.05} & \textbf{22.49} & \textbf{0.39} & \textbf{0.58} \\

            S/2    & focused linear (GELU) & 256 & 256 & 400K & 70.83 & 19.41 & 0.36 & 0.54 \\

            \hline
            B/2    & softmax & 256 & 256 & 400K & 43.47 & -  \\
            
            B/2    & ReLU linear & 256 & 256 & 400K & 56.92 & 25.80 & 0.42 & 0.59 \\

            \gr
            B/2    & Simplified linear (ReLU) & 256 & 256 & 400K & 42.11 & 34.60 & 0.48 & 0.63 \\
            
            B/2    & focused linear (ReLU) & 256 & 256 & 400K & \textbf{40.58} & \textbf{35.98} & \textbf{0.50} & \textbf{0.63} \\

            B/2    & focused linear (GELU) & 256 & 256 & 400K & 58.86 & 24.23 & 0.42 & 0.57 \\
            
            \hline
        \end{tabular}
    \end{center}
    % \vspace{-0.2in}
    \caption{\textbf{Detailed results of Tab. 1 in the main paper.} We report FID-50K~\cite{heusel2017gans} (\textit{without} classifier-free guidance), Inception Score~\cite{salimans2016improved} and Precision/Recall~\cite{kynkaanniemi2019improved} metrics.}
    \label{table-7-1}
    % \vspace{-0.2in}
\end{table*}

\begin{table*}
    \begin{center}
        \small
        \scalebox{1.0}{
        \begin{tabular}{lccccccccc}
            \hline
            DiT & Head  & Resolution & Batch Size & Training Steps   & FID-50K ($\downarrow$) & IS ($\uparrow$) & Precision ($\uparrow$) & Recall ($\uparrow$)	\\
            \hline
            S/2    & 1 & 256 & 256 & 400K & 64.42 & 21.54 & 0.380 & 0.574  \\

            \gr
            S/2    & 2 & 256 & 256 & 400K & 63.24 & 22.07 & 0.385 & 0.570  \\

            S/2    & 3 & 256 & 256 & 400K & \textbf{63.21} & \textbf{22.08} & \textbf{0.386} & \textbf{0.583}  \\

            S/2    & \underline{6} & 256 & 256 & 400K & 63.66 & 22.16 & 0.383 & 0.580  \\

            S/2    & 48 & 256 & 256 & 400K & 78.76 & 17.46 & 0.322 & 0.482  \\

            S/2    & 96 & 256 & 256 & 400K & 116.00 & 11.49 & 0.224 & 0.261  \\

            \hline
            B/2    & 1 & 256 & 256 & 400K & 41.77 & 34.78 & 0.487 & 0.631  \\
 
            B/2    & 2 & 256 & 256 & 400K & 41.39 & 35.59 & 0.494 & 0.631  \\
            
            \gr
            B/2    & 3 & 256 & 256 & 400K & \textbf{40.86} & \textbf{35.79} & \textbf{0.497} & \textbf{0.629} \\

            B/2    & \underline{12} & 256 & 256 & 400K & 42.11 & 34.60 & 0.484 & 0.631  \\

            B/2    & 96 & 256 & 256 & 400K & 68.30 & 20.45 & 0.375 & 0.531  \\

            B/2    & 192 & 256 & 256 & 400K & 112.39 & 12.07 & 0.240 & 0.282  \\
            \hline
            L/2    & 1 & 256 & 256 & 400K & 24.46 & 57.36 & 0.600 & 0.637  \\

            L/2    & 2 & 256 & 256 & 400K & 24.37 & 57.02 & 0.599 & 0.622 \\

            \gr
            L/2    & 4 & 256 & 256 & 400K & \textbf{24.04} & \textbf{59.02} & \textbf{0.597} & \textbf{0.636}  \\

            L/2    & \underline{16} & 256 & 256 & 400K & 25.25 & 54.67 & 0.587 & 0.632  \\

            \hline
            XL/2    & 1 & 256 & 256 & 400K & 21.13 & 65.06 & 0.619 & 0.632  \\

            \gr
            XL/2    & 2 & 256 & 256 & 400K & \textbf{20.66} & \textbf{65.39} & \textbf{0.624} & \textbf{0.636}  \\

            \gr
            XL/2    & 4 & 256 & 256 & 400K & 20.82 & 65.52 & 0.619 & 0.632  \\

            XL/2    & \underline{16} & 256 & 256 & 400K & 21.69 & 63.06 & 0.617 & 0.628  \\

            \hline
        \end{tabular}}
    \end{center}
    % \vspace{-0.2in}
    \caption{\textbf{Detailed results of Tab. 2 in the main paper.} We report FID-50K~\cite{heusel2017gans} (\textit{without} classifier-free guidance), Inception Score~\cite{salimans2016improved} and Precision/Recall~\cite{kynkaanniemi2019improved} metrics. \underline{DiTs~\cite{peebles2023scalable} setting}.}
    \label{table-7-2}
    % \vspace{-0.2in}
\end{table*}

\begin{table*}
    \begin{center}
        \small
        \begin{tabular}{lcccccccc}
            \hline
            Load & Iterations & FFN & Modulation Layer & Attention  & FID-50K ($\downarrow$) & IS ($\uparrow$) & Precision ($\uparrow$) & Recall ($\uparrow$) \\
            \hline
            
            \gr
            model  & 400K & \ding{51} & \ding{51} & \ding{55} & 56.07 & 25.62 & 0.418 & 0.608  \\

            ema  & 400K & \ding{51} & \ding{51} & \ding{55} & 56.07 & 25.61 & 0.416 & 0.601  \\

            \hline

            model  & 200K & \ding{51} & \ding{51} & \ding{55} & 57.84 & 24.72 & 0.408 & 0.600 \\

            model  & 300K & \ding{51} & \ding{51} & \ding{55} & 56.95 & 25.04 & 0.414 & 0.608 \\

            model  & 400K & \ding{51} & \ding{51} & \ding{55} & 56.07 & 25.62 & 0.418 & 0.608 \\

            model  & 600K & \ding{51} & \ding{51} & \ding{55} & 54.80 & 26.65 & 0.424 & 0.613 \\

            \gr
            model  & 800K & \ding{51} & \ding{51} & \ding{55} & \textbf{53.83} & \textbf{27.16} & \textbf{0.425} & \textbf{0.614} \\

            \hline

            model  & 600K & \ding{51} & \ding{51} & Q, K, V & 55.29 & 26.09 & 0.419 & 0.619 \\

            model  & 600K & \ding{51} & \ding{51} & K, V & 55.07 & 26.38 & 0.422 & 0.609 \\

            model  & 600K & \ding{51} & \ding{51} & V & 54.93 & 26.44 & 0.427 & 0.612 \\

            model  & 600K & \ding{51} & \ding{51} & Q & 54.82 & 26.72 & 0.423 & 0.605 \\

            model  & 600K & \ding{51} & \ding{51} & O & 54.84 & 26.33 & 0.425 & 0.607 \\
            
            \hline
        \end{tabular}
    \end{center}
    % \vspace{-0.2in}
    \caption{\textbf{Detailed results of Tab. 3 in the main paper.} We report FID-50K~\cite{heusel2017gans} (\textit{without} classifier-free guidance), Inception Score~\cite{salimans2016improved} and Precision/Recall~\cite{kynkaanniemi2019improved} metrics.}
    \label{table-7-3}
    % \vspace{-0.2in}
\end{table*}

\begin{table*}
    \begin{center}
        \small
        \begin{tabular}{lccccccccc}
            \hline
             Model Size & Iterations & Teacher & $\lambda_1$ & $\lambda_2$ & Training Steps & FID-50K ($\downarrow$) & IS ($\uparrow$) & Precision ($\uparrow$) & Recall ($\uparrow$) \\
            \hline

            S/2 & 800K & DiT-S/2 & 0.1 & 0.0 & 400K & 55.11 & 26.28 & 0.419 & 0.614 \\

            S/2 & 800K & DiT-XL/2 & 0.0 & 0.0 & 400K & \underline{53.83} & \underline{27.16} & \underline{0.425} & \underline{0.614} \\
            
            S/2 & 800K & DiT-XL/2 & 0.1 & 0.0 & 400K & 53.05 & 27.43 & 0.431 & 0.609 \\

            S/2 & 800K & DiT-XL/2 & 0.05 & 0.0 & 400K & 53.41 & 27.26 & 0.427 & 0.610 \\
            
            \gr
            
            S/2 & 800K & DiT-XL/2 & 0.5 & 0.0 & 400K & 51.13 & 28.89 & 0.438 & 0.616 \\

            \hline
            
            S/2 & 800K & DiT-XL/2 & 0.1 & 0.05 & 400K & 52.76 & 27.70 & 0.431 & 0.620 \\

            S/2 & 800K & DiT-XL/2 & 0.0 & 0.05 & 400K & 53.49 & 27.26 & 0.429 & 0.609 \\

            S/2 & 800K & DiT-XL/2 & 0.05 & 0.05 & 400K & 53.14 & 27.46 & 0.431 & 0.609 \\

            \gr
            
            S/2 & 800K & DiT-XL/2 & 0.5 & 0.05 & 400K & \textbf{50.79} & \textbf{29.17} & \textbf{0.443} & \textbf{0.618} \\

            \hline
        \end{tabular}
    \end{center}
    % \vspace{-0.2in}
    \caption{\textbf{Detailed results of Tab. 4 in the main paper.} We report FID-50K~\cite{heusel2017gans} (\textit{without} classifier-free guidance), Inception Score~\cite{salimans2016improved} and Precision/Recall~\cite{kynkaanniemi2019improved} metrics.}
    \label{table-7-4}
    % \vspace{-0.3in}
\end{table*}

\begin{figure*}[t]
	\centering
	\includegraphics[width=1.0\linewidth]{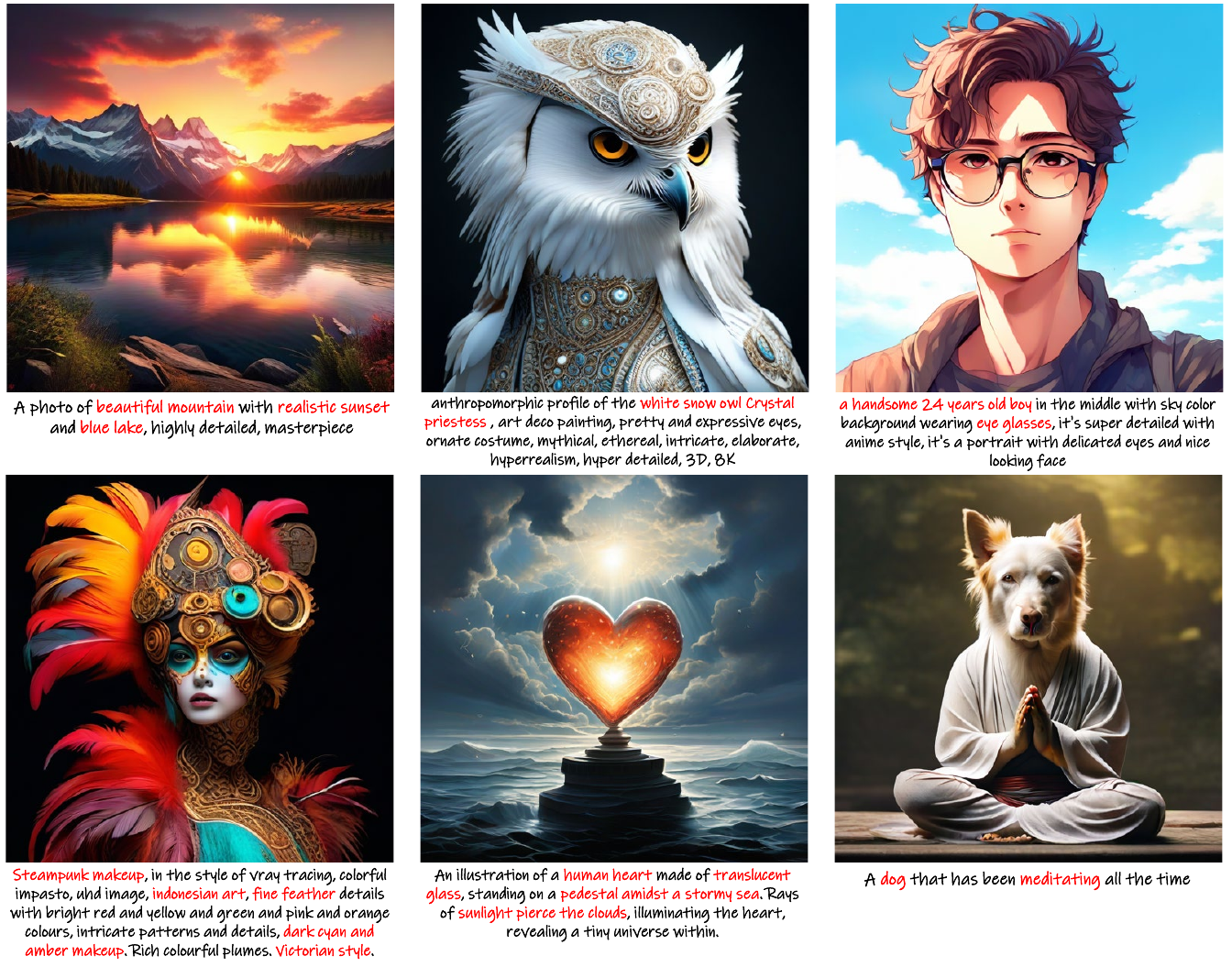}
    \vspace{-0.2in}
	\caption{\textbf{512px Generated samples of LiT following user instructions.} Converted from PixArt-$\Sigma$~\cite{chen2024pixart3}, LiT adopts the same macro- and micro-level architecture, maintaining alignment with the PixArt-$\Sigma$ framework while elegantly replacing all self-attention with efficient linear attention. While being simple and efficient, LiT can generate exceptional high-resolution images following user instructions.}
	\label{fig:t2i-outputs}
    \vspace{-0.2in}
\end{figure*}

\begin{figure*}[t]
	\centering
	\includegraphics[width=1.0\linewidth]{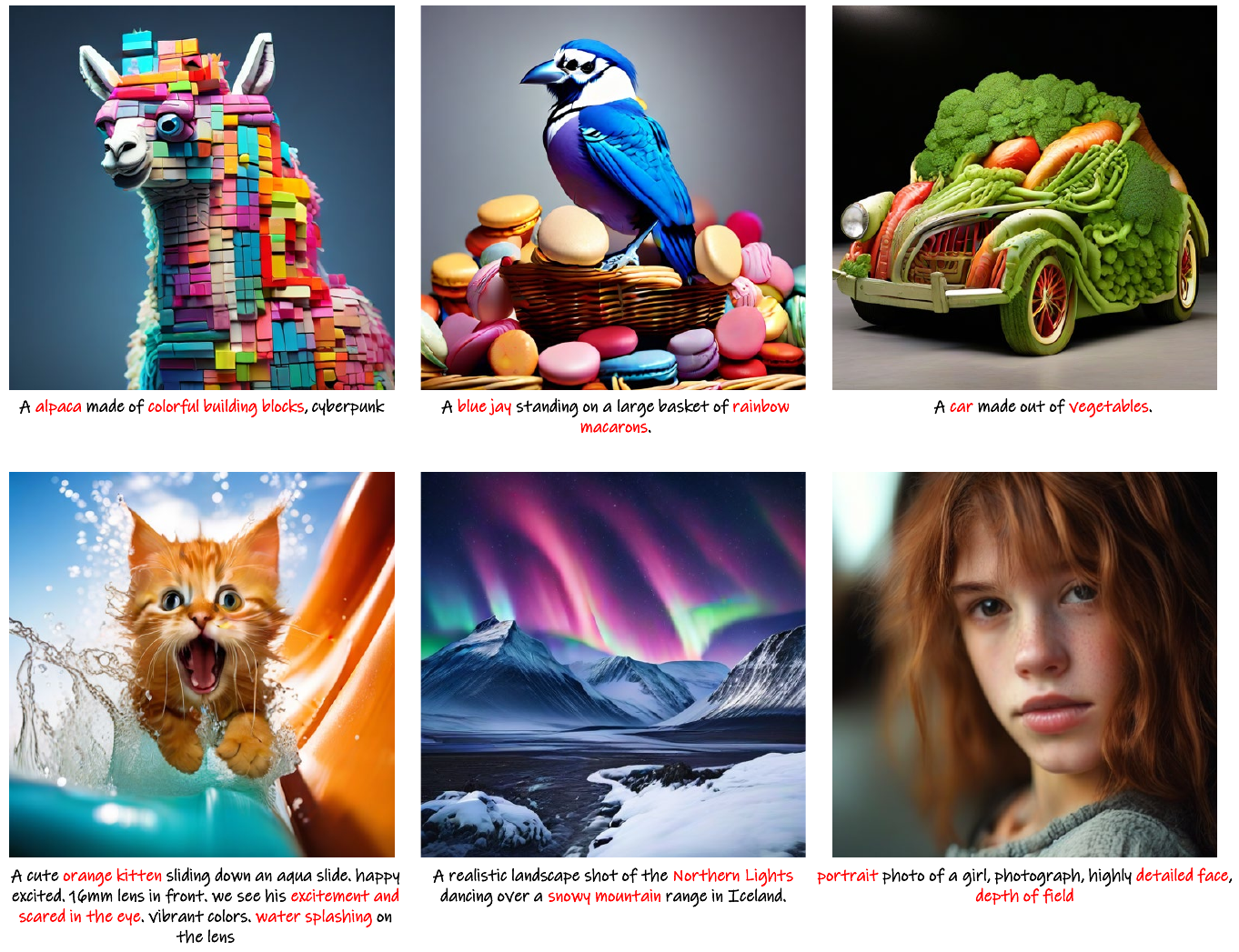}
    % \vspace{-0.2in}
	\caption{\textbf{512px Generated samples of LiT following user instructions.} Converted from PixArt-$\Sigma$~\cite{chen2024pixart3}, LiT adopts the same macro- and micro-level architecture, maintaining alignment with the PixArt-$\Sigma$ framework while elegantly replacing all self-attention with efficient linear attention. While being simple and efficient, LiT can generate exceptional high-resolution images following user instructions.}
	\label{fig:t2i-outputs2}
    % \vspace{-0.2in}
\end{figure*}

\begin{figure*}[t]
	\centering
	\includegraphics[width=1.0\linewidth]{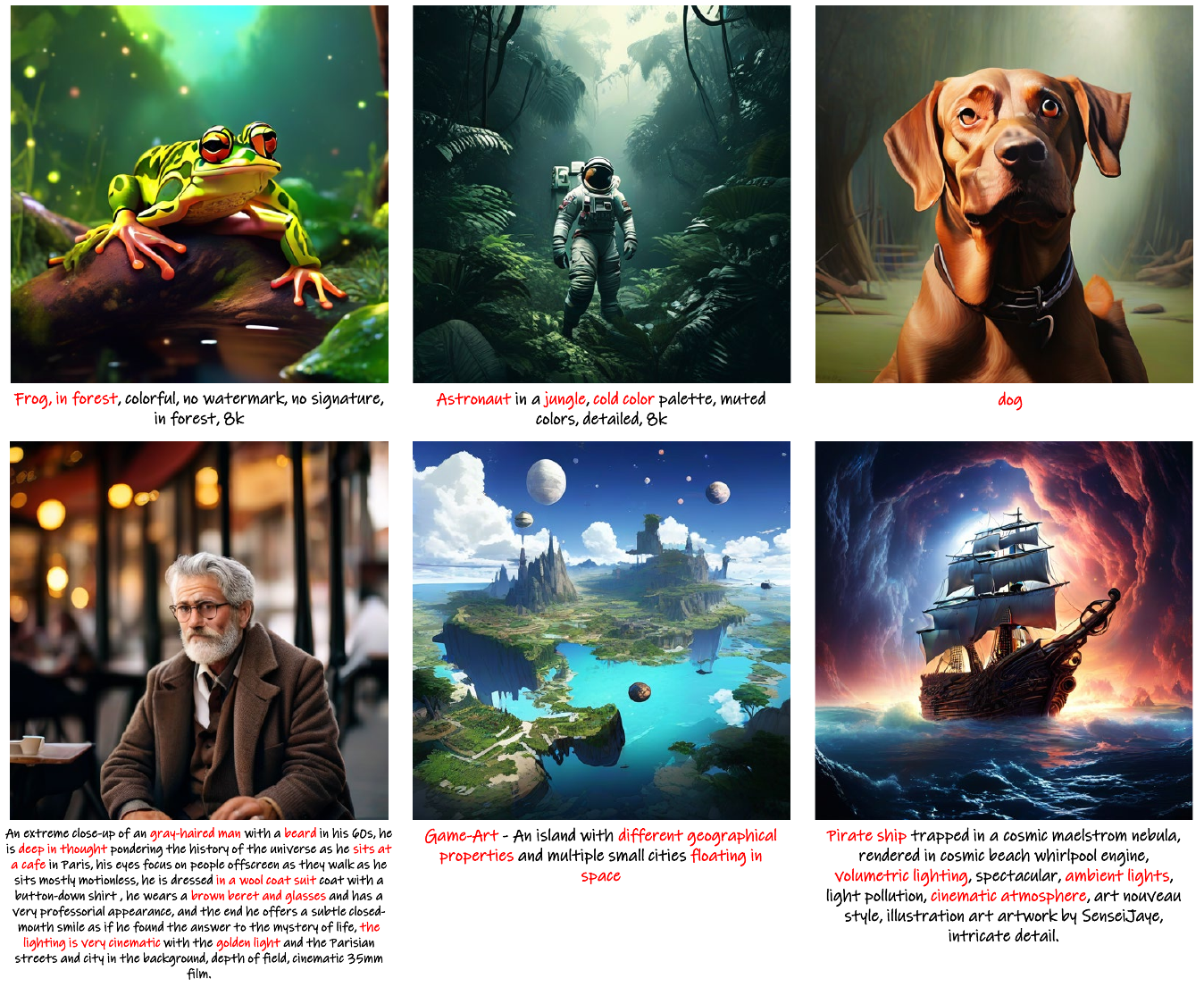}
    % \vspace{-0.2in}
	\caption{\textbf{512px Generated samples of LiT following user instructions.} Converted from PixArt-$\Sigma$~\cite{chen2024pixart3}, LiT adopts the same macro- and micro-level architecture, maintaining alignment with the PixArt-$\Sigma$ framework while elegantly replacing all self-attention with efficient linear attention. While being simple and efficient, LiT can generate exceptional high-resolution images following user instructions.}
	\label{fig:t2i-outputs3}
    % \vspace{-0.2in}
\end{figure*}

\clearpage

\clearpage

\end{document}